\documentclass[11pt,a4paper]{article}
\usepackage[hyperref]{acl2021}
\usepackage{times}
\usepackage{latexsym}
\usepackage{soul, color}

\usepackage{graphicx}

\aclfinalcopy 

\title{Deep Neural Approaches to Relation Triplets Extraction: A~Comprehensive~Survey}

\author{Tapas Nayak$^\dagger$,
  Navonil Majumder$^\diamond$,
  Pawan Goyal$^\dagger$,
  Soujanya Poria$^\diamond$ \\\\
  $^\dagger$ IIT Kharagpur, India \\
  $^\diamond$ Singapore University of Technology and Design, Singapore \\
  \texttt{tnk02.05@gmail.com},\\
  \texttt{\{navonil\_majumder,sporia\}@sutd.edu.sg},\\
  \texttt{pawang@cse.iitkgp.ac.in}}

\date{}

\begin{document}
\maketitle
\begin{abstract}
Recently, with the advances made in continuous representation of words (word embeddings) and deep neural architectures, many research works are published in the area of relation extraction and it is very difficult to keep track of so many papers. To help future research, we present a comprehensive review of the recently published research works in relation extraction. We mostly focus on relation extraction using deep neural networks which have achieved state-of-the-art performance on publicly available datasets. In this survey, we cover sentence-level relation extraction to document-level relation extraction, pipeline-based approaches to joint extraction approaches, annotated datasets to distantly supervised datasets along with few very recent research directions such as zero-shot or few-shot relation extraction, noise mitigation in distantly supervised datasets. Regarding neural architectures, we cover convolutional models, recurrent network models, attention network models, and graph convolutional models in this survey. 
\end{abstract}

\section{Introduction}

A relation triplet consists of two entities and a relation between them. We can find such triplets in a structured format in several publicly available knowledge bases (KBs) such as, Freebase~\cite{bollacker2008freebase}, DBpedia~\cite{bizer2009dbpedia}, Wikidata~\cite{wikidata}, etc. These triplets are very useful for many natural language processing tasks such as machine reading comprehension \cite{qiu2019machine}, machine translation \cite{Zhao2020KnowledgeGE}, abstractive summarization \cite{huang2020KnowledgeGA}, etc. However, building such knowledge bases is a daunting task. The aforementioned KBs are built by crowdsourcing, which may not be scalable. Although these KBs contain a large number of triplets, they remain incomplete. On the other hand, relation triplets can be automatically distilled from the copious amount of free text on the Web. This can be leveraged for identifying missing links in the existing KBs or build a KB from scratch without human intervention. 

\begin{table*}[ht]
\small
\centering
\begin{tabular}{l|l|l}
\hline
\multicolumn{1}{c|}{Class} & \multicolumn{1}{c|}{Sentence}                                                                                                                          & \multicolumn{1}{c}{Triplets}                                                                                                                                                                                 \\ \hline
NEO                                  & \begin{tabular}[c]{@{}l@{}}The original Joy of Cooking was published in 1931 \\by Irma Rombauer, a St. Louis housewife.\end{tabular} & \textless{}Irma Rombauer, St. Louis, place\_lived\textgreater{}                                                                                                                                      \\ \hline
EPO                                     & \begin{tabular}[c]{@{}l@{}}Berlin is the capital of Germany.\end{tabular}                                                                                                                     & \begin{tabular}[c]{@{}l@{}}\textless{}Germany, Berlin, capital\textgreater\\ \textless{}Germany, Berlin, contains\textgreater\\ \textless{}Berlin, Germany, country\textgreater{}\end{tabular} \\ \hline
SEO                                     & \begin{tabular}[c]{@{}l@{}}Dr. C. V. Raman who was born in Chennai worked \\mostly in Kolkata.\end{tabular}                                      & \begin{tabular}[c]{@{}l@{}} \textless{}Dr. C. V. Raman, Chennai, birth\_place\textgreater\\  \textless{}Dr. C. V. Raman, Kolkata, place\_lived\textgreater{}\end{tabular}                         \\ \hline
\end{tabular}
\caption{Examples of different classes of overlapping relation triplets. This table is taken from \newcite{nayak2020deep}.}
\label{tab:ov_classes}
\end{table*}

There are two distinct research paradigms of relation extraction: open information extraction (Open IE) and supervised relation extraction. \newcite{banko2007open}, \newcite{christensen2011srlie}, \newcite{etzioni2011reverb}, and \newcite{schmitz2012ollie} use open information extraction (Open IE) to extract relation triplets from sentences where relations set is open. Open IE systems like KnowItAll \cite{etzioni2004knowitall}, TEXTRUNNER \cite{yates2007textrunner}, REVERB \cite{etzioni2011reverb}, SRLIE \cite{christensen2011srlie}, and OLLIE \cite{schmitz2012ollie} use rule-based methods to extract entities from the noun phrases and relations from the verb phrases present in sentences. These systems can extract a large number of triplets of diverse relations from text within a reasonable time frame. These models extract any verb phrase in the sentences as a relation thus yielding too many uninformative triplets. Also, a relation can be expressed in sentences with different surface forms (lives\_in relation can be expressed with `lives in', `stays', `settles', `lodges', `resident of', etc) and Open IE treats them as different relations which leads to duplication of triplets.

The problems of the Open IE can be addressed using supervised relation extraction. In supervised relation extraction, we consider a fixed set of relations, thus there is no need to do any normalization of the extracted relations. This approach requires a large parallel corpus of text and relation triplets for training. There are some annotated and some distantly supervised parallel corpus of (text, triplets) available publicly that can be used for training the models. Creating annotated corpus is difficult and time-consuming, so datasets created in this way are relatively smaller in size. On the other hand, the distant supervision approach can be exploited to create a large training corpus automatically, but these datasets contain a significant amount of noisy labels. These noisy labels in the distantly supervised datasets can affect the performance of the models in a negative way. Several feature-based models and deep neural network-based are proposed in the last decade for relation extraction. In this survey, we discuss these datasets and models in detail in the remaining part of the paper.

Previously, \newcite{Cui2017ASO,Pawar2017RelationE,Kumar2017ASO,Shi2019ABS,Han2020MoreDM} presented survey of the research works in the relation extraction, but they mostly focused on pipeline-based relation extraction approaches at the sentence-level. Different from these survey papers, we extend the survey to document-level relation extraction and joint entity and relation extraction approaches. We also survey very recent research directions in this area such as zero-shot or few-shot relation extraction and noise mitigation in distantly supervised datasets. To the best of our knowledge, this is the first survey paper that covers so many different aspects of relation extraction in detail.

\begin{table*}[ht]
\small
\centering
\begin{tabular}{cccl}
\hline
\multicolumn{1}{c}{Relation} & \textcolor{red}{Entity 1} & \textcolor{blue}{Entity 2}                                    & \multicolumn{1}{c}{Text}                                                                                                                                                                \\ \hline
acted\_in                        & \textcolor{red}{Meera Jasmine} & \textcolor{blue}{Sootradharan} & \begin{tabular}[c]{@{}l@{}}\textcolor{red}{Meera Jasmine} made her debut in the \\Malayalam film ``\textcolor{blue}{Soothradharan}" .\end{tabular}                                                                         \\ \\ 
located\_in                    & \textcolor{red}{Chakkarakadavu} & \textcolor{blue}{Kerala}     & \begin{tabular}[c]{@{}l@{}}\textcolor{red}{Chakkarakadavu} is a small village to \\the east of the town of Cherai, on \\Vypin Island in Ernakulam district,\\  \textcolor{blue}{Kerala}, India .\end{tabular}             \\ \\
birth\_place                      & \textcolor{red}{Barack Obama} & \textcolor{blue}{Hawaii}  & \begin{tabular}[c]{@{}l@{}}\textcolor{red}{Barack Obama} was born in \textcolor{blue}{Hawaii} . \end{tabular} \\ \\
plays\_for                       & \textcolor{red}{Moussa Sylla} & \textcolor{blue}{Horoya AC}   & \begin{tabular}[c]{@{}l@{}}Fodé \textcolor{red}{Moussa Sylla} is a Guinean \\football player, who currently plays \\for \textcolor{blue}{Horoya AC} .\end{tabular}                      \\ \\
owns                           & \textcolor{red}{MTV Channel} & \textcolor{blue}{Shakthi TV}    & \begin{tabular}[c]{@{}l@{}}\textcolor{red}{MTV Channel} (Pvt) Ltd is a Sri \\Lankan media company which owns \\three national television channels - \\\textcolor{blue}{Shakthi TV}, Sirasa TV and TV 1 .\end{tabular} \\ \hline
\end{tabular}
\caption{Examples of relation triplets found in free texts. This table is taken from \newcite{nayak2020deep}.}
\label{tab:text_triplet}
\end{table*}

\section{Task Description}

Given a sentence and a set of relations $R$ as input, the task is to extract a set of relation triplets, with relations from $R$, from the sentence. Pipeline-based relation extraction approaches divide the task into two sub-tasks: (i) entity recognition and (ii) relation classification. In the first sub-task, all the candidate entities are identified in a sentence. In the second sub-task, the relation between every possible ordered pair of candidate entities is determined --- this relation may not exist ({\em None}). 

Joint-extraction approaches, in contrast, jointly find the entities and relations. Joint models extract only the valid relational triplets and they do not need to extract the {\em None} triplets. Relation triplets may share one or both entities among them and this overlapping of entities makes this task challenging. Based on the overlap of entities, we divide the sentences into three classes: (i) {\em No Entity Overlap (NEO)}: A sentence in this class has one or more triplets, but they do not share any entities. (ii) {\em Entity Pair Overlap (EPO)}: A sentence in this class has more than one triplet, and at least two triplets share both the entities in the same or reverse order. (iii) {\em Single Entity Overlap (SEO)}: A sentence in this class has more than one triplet and at least two triplets share exactly one entity. It should be noted that a sentence can belong to both EPO and SEO classes. The goal is to extract all relation triplets present in a sentence.

\section{Scope of this Survey}

In this survey, we focus on the relation triplets concerning PERSON, ORGANIZATION, and LOCATION mainly. Many research works are published for domain-specific relation extraction such scientific articles \cite{luan2017scientific,jain2020scirex}, medical \cite{Gu2016ChemicalinducedDR,Li2017ANJ,Choi2018ExtractionOP,Thillaisundaram2019BiomedicalRE}, legal \cite{andrew2018automatic}, finance \cite{Vela2009ConceptAR}, etc. But in this survey, we do not include the research papers that only focus on a particular domain. Also, we only focus on relation extraction for the English language.

\section{Challenges of Dataset Annotation}

Existing KBs, such as Freebase, Wikidata, and DBpedia, are manually built which takes much effort and time. However, these KBs still have a large number of missing links. On the other hand, we can find evidence of a large number of relation triplets in free texts. We have included some examples of such triplets and texts in Table \ref{tab:text_triplet}. If we can extract relation triplets automatically from the text, we can build a KB from scratch or add new triplets to the existing KBs without any manual effort. But to achieve this goal, we need a large number of texts annotated with relation triplets, and creating such a corpus manually is a daunting task. One possible way to do the annotation is to identify the entities in the text and then for all possible pairs of entities, identify the relations from a pre-defined set of relations or {\em None} if none of the relations from this set holds in this text. The identification of the entities in a text is relatively easier, but the difficulty of identifying the relations from a set grows with the size of the relations set. For few relations such as 3/4/5, this may be easier, but when the number of relations grows to 20/30/40, it becomes very challenging. Marking the {\em None} relations in the case of large relations set is more difficult as the annotators have to make sure that none of the relations from the set holds between two entities in the text.

\begin{table*}[ht]
\small
\centering
\begin{tabular}{lccccc}
\hline
\multicolumn{1}{c}{Text}                                                                                           & \textcolor{red}{Entity 1}                                                & \textcolor{blue}{Entity 2}                                               & \begin{tabular}[c]{@{}c@{}}Distantly \\ Supervised \\ Relation\end{tabular} & \begin{tabular}[c]{@{}c@{}}Actual\\ Relation\end{tabular} & Status \\ \hline
\begin{tabular}[c]{@{}l@{}}\textcolor{red}{Barack Obama} was born in \textcolor{blue}{Hawaii} .\end{tabular}                                             & \begin{tabular}[c]{@{}c@{}}\textcolor{red}{Barack} \\ \textcolor{red}{Obama}\end{tabular} & \textcolor{blue}{Hawaii}                                                 & birth\_place                                                          & birth\_place                                          & Clean         \\ \\
\begin{tabular}[c]{@{}l@{}}\textcolor{red}{Barack Obama} visited \textcolor{blue}{Hawaii} .\end{tabular}                                                 & \begin{tabular}[c]{@{}c@{}}\textcolor{red}{Barack} \\ \textcolor{red}{Obama}\end{tabular} & \textcolor{blue}{Hawaii}                                                 & birth\_place                                                          & None                                                      & Noisy        \\ \\
\begin{tabular}[c]{@{}l@{}}Suvendu Adhikari was born at \textcolor{red}{Karkuli} \\in Purba Medinipur in \textcolor{blue}{West Bengal} .\end{tabular} & \textcolor{red}{Karkuli}                                                 & \begin{tabular}[c]{@{}c@{}}\textcolor{blue}{West}  \textcolor{blue}{Bengal}\end{tabular} & None                                                                      & located\_in                                           & Noisy        \\ \\
\begin{tabular}[c]{@{}l@{}}Suvendu Adhikari, transport minister \\of \textcolor{blue}{West Bengal}, visited \textcolor{red}{Karkuli} .\end{tabular}  & \textcolor{red}{Karkuli}                                                 & \begin{tabular}[c]{@{}c@{}}\textcolor{blue}{West} \\ \textcolor{blue}{Bengal}\end{tabular} & None                                                                      & None                                                      & Clean         \\ \hline
\end{tabular}
\caption{Examples of distantly supervised clean and noisy samples. This table is taken from \newcite{nayak2020deep}.}
\label{tab:noisy_examples}
\end{table*}

To overcome the dataset annotation problems, \newcite{mintz2009distant,riedel2010modeling,hoffmann2011knowledge} proposed the idea of distant supervision to automatically obtain the text-triplet mapping without any human effort. In distant supervision, the triplets from an existing KB are mapped to a free text corpus such as Wikipedia articles or news articles (e.g., New York Times). The idea of distant supervision is that if a sentence contains two entities of a triplet from a KB, that sentence can be considered as the source of this KB triplet. On the other hand, if a sentence contains two entities from a KB and there is no relation between these two entities in the KB, that sentence is considered as a source of {\em None} triplet between the two entities. These {\em None} samples are useful as distantly supervised models consider only a limited set of positive relations. Any relation outside this set is considered as {\em None} relation. This method can give us a large number of triplet-to-text mappings which can be used to build supervised models for this task. This idea of distant supervision can be extended easily to single-document or multi-document relation extraction.

But the distantly supervised data may contain many noisy samples. Sometimes, a sentence may contain the two entities of a positive triplet, but the sentence may not express any relation between them. These kinds of sentences and entity pairs are considered as noisy positive samples. Another set of noisy samples comes from the way samples for {\em None} relation are created. If a sentence contains two entities from the KB and there is no relation between these two entities in the KB, this sentence and entity pair is considered as a sample for {\em None} relation. But knowledge bases are often not complete and many valid relations between entities in a KB are missing. So it may be possible that the sentence contains information about some positive relation between the two entities, but since that relation is not present in the KB, this sentence and entity pair is incorrectly considered as a sample for {\em None} relation. These kinds of sentences and entity pairs are considered as noisy negative samples.

We include examples of clean and noisy samples generated using distant supervision in Table \ref{tab:noisy_examples}. The KB contains many entities out of which four entities are {\em Barack Obama}, {\em Hawaii}, {\em Karkuli}, and {\em West Bengal}. {\em Barack Obama} and {\em Hawaii} have a {\em birth\_place} relation between them. Karkuli and West Bengal are not connected by any relations in the KB. So we assume that there is no valid relation between these two entities. The sentence in the first sample contains the two entities {\em Barack Obama} and {\em Hawaii}, and it also contains information about {\em Obama} being born in {\em Hawaii}. So this sentence is a correct source for the triplet ({\em Barack Obama}, {\em Hawaii}, {\em birth\_place}). So this is a clean positive sample. The sentence in the second sample contains the two entities, but it does not contain the information about {\em Barack Obama} being born in {\em Hawaii}. So it is a noisy positive sample. In the case of the third and fourth samples, according to distant supervision, they are considered as samples for {\em None} relation. But the sentence in the third sample contains the information for the actual relation {\em located\_in} between {\em Karkuli} and {\em West Bengal}, even though the KB happens not to contain the {\em located\_in} relation relating {\em Karkuli} and {\em West Bengal}. So the third sample is a noisy negative sample. The fourth sample is an example of a clean negative sample.

Despite the presence of noisy samples, relation extraction models trained on distantly supervised data have proven to be successful for relation extraction. These models can be used to fill the missing facts of a KB by automatically finding triplets from free texts. It can save much manual effort towards completing an existing KB. 

\section{Relation Extraction Datasets}

\begin{table*}[ht]
\small
\centering
\begin{tabular}{lccccc}
\hline
Dataset Name & Level & \# Valid Relations & \# Train & \# Test & Manual Annotation \\ \hline
SemEval 2010 Task 8        & sentence         &  18           & 8,000         & 2,717        &  Yes                 \\ 
NYT10        & sentence         &  52           & 455,412         & 172,415        &  No                 \\ 
NYT11        & sentence         &  24           & 335,843         &  1,450       & Test                   \\ 
NYT29        & sentence         &  29           & 63,306         & 4,006        &  No                 \\ 
NYT24        & sentence         &  24           & 56,196         &  5,000       & No                   \\ 
WebNLG        & sentence         &  216           & 5,519         &  703       & Yes                   \\ 
ACE05        & sentence         &  7           &  9,038        &  1,535       & Yes                   \\ 
CoNLL04      & sentence         &  5           &  1,153        &  288       &   Yes                \\ 
GDS          & sentence         &  4           &  13,161        & 5,663        &   Yes                \\ 
TACRED   & sentence         &  41           &  90,755        &  15,509       &   Yes                \\ 
FewRel 2.0       &  sentence        &  100           &  56,000        &  14,000       &  Yes                 \\ 
WikiReading   & document         &  884           &  14.85M        &  3.73M       & No                   \\ 
DocRED   & document        &  96           &  4,053        & 1,000        &  Yes                 \\ \hline
\end{tabular}
\caption{The details of relation extraction datasets.}
\label{tab:re_datasets}
\end{table*}

Several datasets are available for the relation extraction task. \citet{hendrickx2010semeval} proposed a shared task on relation extraction in SemEval 2010 and released a dataset with 8,000 training sentences and 2,717 test instances across nine relations including {\em None}. The relations in this dataset are not taken from any knowledge base. They represent the relationship between two nominals in the sentences. Examples of such relations are {\em Cause-Effect, Component-Whole,} etc. \citet{mintz2009distant} mapped Freebase \citep{bollacker2008freebase} triplets to Wikipedia articles to obtain a dataset. \citet{riedel2010modeling} (NYT10) and \citet{hoffmann2011knowledge} (NYT11) mapped Freebase triplets to the New York Times (NYT) articles to obtain a similar dataset. These two datasets are used extensively by researchers for their experiments. They have $52$ and $24$ valid relations respectively. The training and test data in NYT10 are distantly supervised, whereas in NYT11, the test data is annotated and training data is distantly supervised. Recently, \newcite{zhu2020nyth} created an annotated test dataset for the NYT10 dataset with a subset of its relations set. This annotated test set contains 22 relations. They used a binary strategy to annotate each instance either the distantly supervised relation is present or not in the sentences. But this test dataset does not include any {\em None} samples which makes it unsuitable for the relation extraction task. ACE04 \cite{doddington2004automatic} and ACE05 \cite{ace2005walker} are two datasets containing 7 relations. These two datasets focus on both named entity recognition and relation extraction tasks. CoNLL04 \citep{roth2004linear} and GDS \citep{jat2018attention} are two other datasets with $5$ and $4$ valid relations respectively. ACE04, ACE05, CoNLL04, and GDS datasets are manually annotated but they contain few relations in comparison to distantly supervised datasets. TACRED \cite{zhang2017position} is another dataset for relation extraction that has manually annotated training and test data. TACRED contains 41 relations similar to that of the distantly supervised datasets. So that makes this dataset very suitable for comparing models in this task. Automatic evaluation of the models can be carried out on this dataset easily. FewRel 2.0 \citep{gao2019fewrel} is a few-shot relation extraction dataset. WebNLG \cite{zeng2018copyre} is another dataset that contains $216$ relations. Recently, this dataset has been used for joint entity and relation extraction. It is curated from the original WebNLG dataset of \newcite{Gardent2017CreatingTC}. NYT24 \cite{zeng2018copyre,nayak2019ptrnetdecoding} and NYT29 \cite{takanobu2019hrlre,nayak2019ptrnetdecoding} are two other popular datasets for joint extraction task. These two datasets are curated from the NYT11 and NYT10 datasets respectively after removing the sentences that do not contain any valid relation triplets. These datasets are created at the sentence level. 

WikiReading \citep{hewlett2016wikireading} and DocRED \citep{yao2019DocRED} are two document-level relation extraction datasets created using Wikipedia articles and Wikidata items. WikiReading is a slot-filling dataset where a document of an entity and the name of a property (same as the relation) is given to the models as input to predict the second entity. This dataset does not have any {\em None} instances. Each document in the dataset corresponds to one instance of training or testing. DocRED, on the other hand, is a relation extraction dataset. Training data contains 4,053 documents and test data contains 1,000 documents. Each document contains multiple instances and test data is blind. \newcite{nayak2020deep} proposed an idea of extending the relation extraction task to multi-documents. They created a 2-hop relation extraction dataset from a multi-hop question answering dataset WikiHop \cite{welbl2018constructing} that contains more relations than the previous sentence-level or document-level datasets. Their idea can be extended to create an N-hop dataset to cover more relations. The details of these datasets are included in Table \ref{tab:re_datasets}.

\section{Evaluation Metrics}

In the pipeline approach, the assumption is that entities are already identified and models need to classify the relation or no relation ({\em None}) between the pairs of entities. There are two ways in which the performance of models can be measured: (i) At sentence-level (ii) At bag-level. In the case of the sentence-level, each sentence with an entity pair is considered as a test instance. At the bag-level, a bag of sentences where each sentence must contain the same entity pair is considered as a test instance. In both ways, models are evaluated using precision, recall, and F1 scores after removing the {\em None} labels. A confidence threshold is used to decide if the relation of a test instance belongs to the set of relations $R$ or {\em None}. If the model predicts {\em None} for a test instance, then it is considered as {\em None} only. But if the network predicts a relation from the set $R$ and the corresponding softmax score is below the confidence threshold, then the final prediction label is changed to {\em None}. This confidence threshold is the one that achieves the highest F1 score on the validation dataset. Since most of the test datasets in this task are distantly supervised and they contain noisy samples, automatic evaluation metric such as the F1 score may not be suitable. The precision-recall curve is a popular automatic metric for the evaluation of distantly supervised test datasets. The area under the precision-recall curve (AUC) indicates the performance measure of the models. Precision@K is another metric used for evaluation on such test datasets, but it requires manual effort. 

For the joint extraction approaches, models are evaluated based on the number of the correct triplets extracted from the sentences. The extracted triplets are considered as a set and duplicate triplets are removed. An extracted triplet is considered correct if the corresponding entity names are correct and the relation is also correct. Precision, recall, and F1 scores are measured based on that. There are two variants of matching the entity names. The first one is partial matching (P) where only the last token of the entity names is matched. The second one is exact matching (E) where the full entity names are matched. 

\section{Relation Extraction Models}

Relation extraction models can be categorized into two sets: (i) pipeline extraction approaches (ii) joint extraction approaches. We have included several state-of-the-art models of both the categories below. 

\subsection{Pipeline Extraction Approaches}

At the beginning of relation extraction research, pipeline approaches were quite popular. A pipeline approach has two steps: (i) First, a named entity recognizer is used to identify the named entities in a text. (ii) Next, a classification model is used to find the relation between a pair of entities. The named entities identified in the first step are mapped to the KB entities. There are several state-of-the-art NER models available as proposed by \newcite{Huang2015BidirectionalLM,Ma2016EndtoendSL,lample2016neural,chiu2016named} can be used for this purpose. Contextualized word embeddings based model such as ELMo \cite{peters2018deep}, BERT \cite{devlin2019bert}, RoBERTa \cite{Liu2019RoBERTaAR}, and SpanBERT \cite{Joshi2019SpanBERT} can also be used for named entity recognition. In the next step, different classification models are proposed to find the relations between entity pairs and we describe them in detail in the following subsections.

\subsubsection{Feature-Based Models}

\citet{mintz2009distant} proposed a feature-based relation classification model for this task. They used lexical features such as the sequence of words between two entities and their part-of-speech (POS) tags, a flag indicating which entity appears first, $k$ tokens to the left of entity 1 and $k$ tokens to the right of entity 2, syntactic features such as dependency path between two entities, and named entity types of the two entities in their model. \citet{riedel2010modeling} proposed multi-instance learning for this task to mitigate the problem of noisy sentences obtained using the distant supervision method. They used a factor graph to explicitly model the decision of whether two entities are related and whether this relation is mentioned in a given sentence. Also, they applied constraint-driven semi-supervision to train their model without any knowledge about which sentences express the relations. Their multi-instance learning model significantly improves the performance over the model proposed by \citet{mintz2009distant}. 

\citet{hoffmann2011knowledge} and \citet{mimlre} proposed the idea of multi-instance multi-labels (MIML) to solve the problem of overlapping relations. They used probabilistic graphical models that take a bag of sentences containing two entities as input and find all possible relations between them. Similarly, \citet{ren2017cotype} used a feature-based model to jointly predict the relation between two entities and their fine-grained types. They used features like the head tokens of two entities, tokens of two entities, tokens between the two entities, their POS tags, ordering of the two entities, the distance between them, and the Brown cluster\footnote{\url{https://github.com/percyliang/brown-cluster}} of each token in their model. They proposed a joint optimization framework to learn the entity embeddings, relation embeddings, and fine-grained type embeddings of the entities together.

\subsubsection{CNN-Based Neural Models}

Distributed representations of words as word embeddings have transformed the way that natural language processing tasks like IE can be tackled. Word2Vec \citep{mikolov2013distributed} and GloVe \citep{pennington2014glove} are two sets of large and publicly available word embeddings that are used for many NLP tasks. Most neural network-based models for information extraction have used the distributed representation of words as their core component. The high dimensional distributed representation of words can encode important semantic information about words, which is very helpful for identifying the relations among the entities present in a sentence. Initially, neural models also follow the pipeline approach to solve this task.

\citet{zeng2014relation} used a convolutional neural network for relation extraction. They used the pre-trained word embeddings of \citet{turian2010word} to represent the tokens in a sentence and used two distance embedding vectors to represent the distance of each word from the two entities. They used a convolutional neural network (CNN) and max-pooling operation to extract a sentence-level feature vector. This sentence representation is passed to a feed-forward neural network with a softmax activation layer to classify the relation. 

\begin{figure}[ht]
\centering
\includegraphics[scale=0.2]{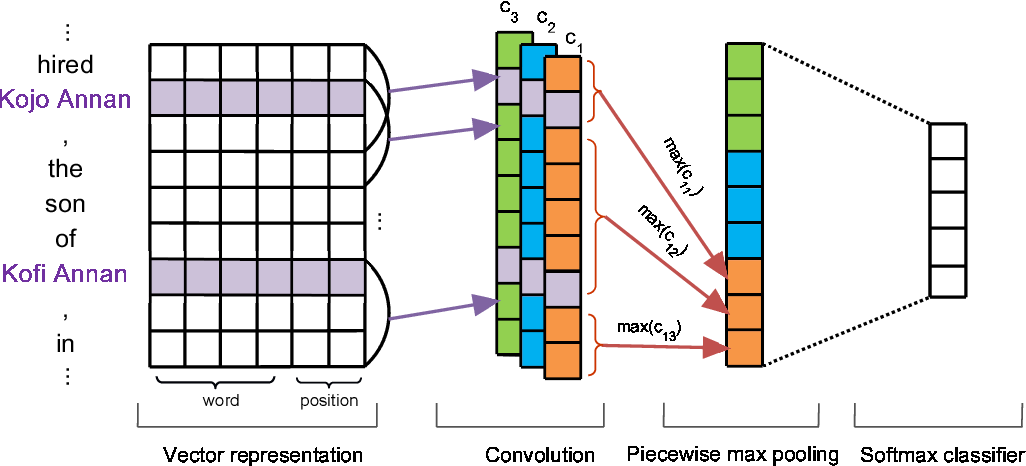}
\caption{The architecture of the PCNN model \cite{zeng2015distant}.}
\label{fig:pcnn}
\end{figure}

\citet{zeng2015distant} introduced a piecewise convolutional neural network (PCNN) to improve relation extraction. \citet{zeng2014relation} applied the max-pooling operation across the entire sentence to get the single important feature from the entire sentence for a particular convolutional filter. In PCNN, the max-pooling operation is not performed for the entire sentence. Instead, the sentence is divided into three segments: from the beginning to the argument appearing first in the sentence, from the argument appearing first in the sentence to the argument appearing second in the sentence, and from the argument appearing second in the sentence to the end of the sentence. Max-pooling is performed in each of these three segments and for each convolutional filter to obtain three feature values. A sentence-level feature vector is obtained by concatenating all such feature values and is given to a feed-forward neural network with a softmax activation layer to classify the relation.

\subsubsection{Attention-Based Neural Models}

Recently, attention networks have proven very useful for different NLP tasks. \citet{huang2016attention}, \newcite{wang2016relation}, \citet{zhang2017position}, and \citet{jat2018attention} used word-level attention model for single-instance sentence-level relation extraction. \citet{huang2016attention} proposed a combination of a convolutional neural network model and an attention network. First, a convolution operation with max-pooling is used to extract the global features of the sentence. Next, attention is applied to the words of the sentence based on the two entities separately. The word embedding of the last token of an entity is concatenated with the embedding of every word. This concatenated representation is passed to a feed-forward layer with tanh activation and then another feed-forward layer with softmax to get a scalar attention score for every word of that entity. The word embeddings are averaged based on the attention scores to get the attentive feature vectors. The global feature vector and two attentive feature vectors for the two entities are concatenated and passed to a feed-forward layer with softmax to determine the relation.

\newcite{wang2016relation} used multi-level attention CNNs for this task. Their model achived very high F1 score on the SemEval 2010 Task 8 dataset. \newcite{zhang2017position} proposed a position-aware attention mechanism over the LSTM sequences for this task. Earlier \newcite{zeng2014relation} and \newcite{zeng2015distant} use the position information as dense embedding in the network for feature extraction, whereas \newcite{zhang2017position} used it in attention modeling for the same task.

\begin{figure}[ht]
\centering
\includegraphics[scale=0.2]{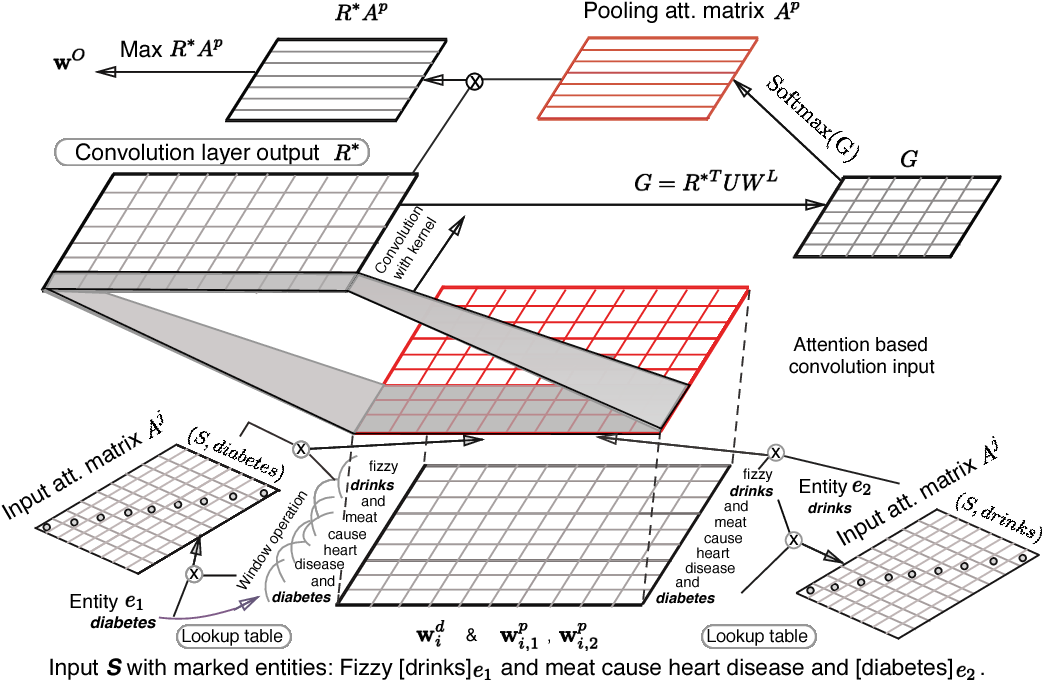}
\caption{The architecture of the multi-level attention CNN model \cite{wang2016relation}.}
\label{fig:multi_att_cnn}
\end{figure}

\citet{jat2018attention} used a bidirectional gated recurrent unit (Bi-GRU) \citep{cho2014properties} to capture the long-term dependency among the words in the sentence. The tokens vectors $\mathbf{x}_t$ are passed to a Bi-GRU layer. The hidden vectors of the Bi-GRU layer are passed to a bi-linear operator which is a combination of two feed-forward layers with softmax to compute a scalar attention score for each word. The hidden vectors of the Bi-GRU layer are multiplied by their corresponding attention scores for scaling up the hidden vectors. A piecewise convolution neural network \citep{zeng2015distant} is applied to the scaled hidden vectors to obtain the feature vector. This feature vector is passed to a feed-forward layer with softmax to determine the relation. \newcite{nayak2019effective} used dependency distance based multi-focused attention model for this task. Dependency distance helps to identify the important words in the sentences and multi-factor attention helps to focus on multiple pieces of evidence for a relation. \newcite{Bowen2019BeyondWA} used segment-level attention in their model rather than using traditional token-level attention for this task. \newcite{Zhang2019MultilabeledRE} proposed an attention-based capsule network for relation extraction.

\citet{lin2016neural} have used attention model for multi-instance relation extraction. They applied attention over a bag of independent sentences containing two entities to extract the relation between them. First, CNN-based models are used to encode the sentences in a bag. Then a bi-linear attention layer is used to determine the importance of each sentence in the bag. This attention helps to mitigate the problem of noisy samples obtained by distant supervision to some extent. The idea is that clean sentences get higher attention scores over the noisy ones. The sentence vectors in the bag are merged in a weighted average fashion based on their attention scores. The weighted average vector of the sentences is passed to a feed-forward neural network with softmax to determine the relation. This bag-level attention is used only for positive relations and not used for {\em None} relation. The reason is that the representations of the bags that express no relations are always diverse and it is difficult to calculate suitable weights for them.

\citet{ye2019distant} used intra-bag and inter-bag attention networks in a multi-instance setting for relation extraction. Their intra-bag attention is similar to the attention used by \citet{lin2016neural}. Additionally, they used inter-bag attention to address the noisy bag problem. They divide the bags belonging to a relation into multiple groups. The attention score for each bag in a group is obtained based on the similarity of the bags to each other within the group. This inter-bag attention is used only during training as we do not know the relations during testing. Similarly \newcite{Yuan2019CrossrelationCA} proposed a cross-relation and cross-bag attention for multi-instance relation extraction. \newcite{Li2020SelfAttentionES} proposed an entity-aware embeddings and self-attention \cite{vaswani2017attention} enhanced PCNN model for relation extraction.

\subsubsection{Dependency-Based Neural Models}

Some previous works have incorporated the dependency structure information of sentences in their neural models for relation extraction. \citet{xu2015classifying} used a long short-term memory network (LSTM) \citep{hochreiter1997long} along the shortest dependency path (SDP) between two entities to find the relation between them. Each token along the SDP is represented using four embeddings -- pre-trained word vector, POS tag embedding, embedding for the dependency relation between the token and its child in the SDP, and embedding for its WordNet \citep{Fellbaum2000WordNetA} hypernym. They divide the SDP into two sub-paths: (i) The left SDP which goes from entity 1 to the common ancestor node (ii) The right SDP which goes from entity 2 to the common ancestor node. This common ancestor node is the lowest common ancestor between the two entities in the dependency tree. The token vectors along the left SDP and right SDP are passed to an LSTM layer separately. A pooling layer is applied to the hidden vectors to extract the feature vector from the left SDP and right SDP. These two vectors are concatenated and passed to a classifier to find the relation. 

\citet{liu2015dependency} exploited the shortest dependency path (SDP) between two entities and the sub-trees attached to that path (augmented dependency path) for relation extraction. Each token in the SDP is represented using its pre-trained embedding and its sub-tree representation. The sub-tree representation of a token is obtained from the sub-tree of the dependency tree where the token is the root node. The dependency relations are represented using trainable embeddings. Each node in the sub-tree of a token receives information from its children including the dependency relations. The sub-tree representation of the token is obtained by following the sub-tree rooted at the token from its leaf nodes to the root in a bottom-up fashion. Next, they use CNN with max-pooling on the vectors of the sequence of the tokens and dependency relations across the SDP. The output of the max-pooling operation is passed to a classifier to find the relation.

\begin{figure}[ht]
\centering
\includegraphics[scale=0.2]{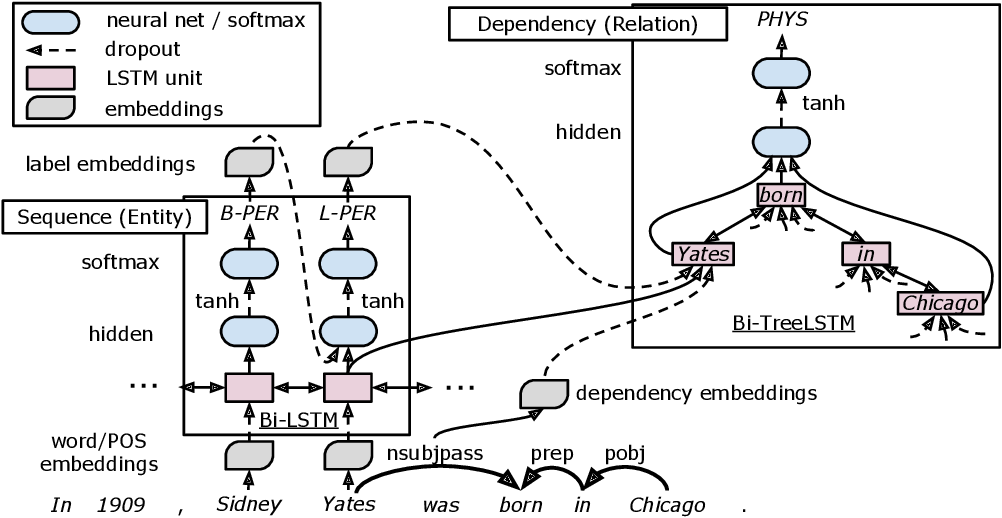}
\caption{The architecture of the relation extraction model using LSTMs on sequences and tree structures \cite{miwa2016end}.}
\label{fig:treelstm}
\end{figure}

\citet{miwa2016end} used a tree LSTM network along the shortest dependency path (SDP) between two entities to find the relation between them. They used a bottom-up tree LSTM and top-down tree LSTM in their model. In the bottom-up tree LSTM, each node receives information from all of its children. The hidden representation of the root node of this bottom-up tree LSTM is used as the final output. In the top-down tree LSTM, each node receives the information from its parent node. The hidden representations of the head token of two entities are the final output of this tree LSTM. The representations of the bottom-up tree LSTM and top-down tree LSTM are concatenated and passed to a classifier to find the relation. They showed that using the SDP tree over the full dependency tree is helpful as unimportant tokens for the relation are ignored in the process. \newcite{Veyseh2020ExploitingTS} proposed a ON-LSTM \cite{Shen2019OrderedNI} based relation extraction model to preserve the syntax consistency in the model.

\subsubsection{Graph-Based Neural Models}

Graph-based models are popular for many NLP tasks as they work on non-linear structures. \citet{Quirk2017DistantSF} proposed a graph-based model for cross-sentence relation extraction. They built a graph from the sentences where every word is considered as a node in the graph. Edges are created based on the adjacency of the words, dependency tree relations, and discourse relations. They extract all the paths from the graph starting from entity 1 to entity 2. Each path is represented by features such as lexical tokens, the lemma of the tokens, POS tags, etc. They use all the path features to find the relation between the two entities. 

\citet{peng2017cross} and \citet{Song2018NaryRE} used a similar graph for N-ary cross-sentence relation extraction. Rather than using explicit paths, they used an LSTM on a graph. A graph LSTM is a general structure for a linear LSTM or tree LSTM. If the graph contains only the word adjacency edges, then the graph LSTM becomes a linear LSTM. If the graph contains the edges from the dependency tree, it becomes a tree LSTM. A general graph structure may contain cycles. So \citet{peng2017cross} divides this graph into two directed acyclic graphs (DAG), where the forward DAG contains only the forward edges among the tokens and the backward DAG contains only the backward edges among the tokens. Each node has a separate forget gate for each of its neighbors. It receives information from the neighbors and updates its hidden states using LSTM equations \citep{hochreiter1997long}. If we only consider the word adjacency edges, this graph LSTM becomes a bi-directional linear LSTM. \citet{Song2018NaryRE} did not divide the graph into two DAGs, but directly used the graph structure to update the states of the nodes. At time step $t$, each node receives information from its neighbor from the previous time step and update its hidden states using LSTM equations. This process is repeated $k$ number of times where $k$ is a hyper-parameter.

\citet{Kipf2017SemiSupervisedCW} and \citet{velickovic2018graph} proposed a graph convolutional network (GCN) model which used simple linear transformations to update the node states, unlike the graph LSTMs used by \citet{peng2017cross} and \citet{Song2018NaryRE}. \citet{Kipf2017SemiSupervisedCW} gave equal weights to the edges, whereas \citet{velickovic2018graph} used an attention mechanism to assign different weights to the edges. \citet{vashishth2018reside}, \citet{zhang2018graph}, and \citet{guo2019aggcn} used graph convolutional networks for sentence-level relation extraction. They considered each token in a sentence as a node in the graph and used the syntactic dependency tree to create a graph structure among the nodes. \citet{vashishth2018reside} used the GCN in a multi-instance setting. They used a Bi-GRU layer and a GCN layer over the full dependency tree of the sentences to encode them. The sentence representations in a bag were aggregated and passed to a classifier to find the relation. Following \citet{miwa2016end}, \citet{zhang2018graph} used only the shortest dependency path (SDP) tree to build the adjacency matrix for the graph. Along with the SDP tree, they included the edges that are distance $K$ away from the SDP where $K$ is a hyper-parameter. \citet{guo2019aggcn} proposed a soft pruning strategy over the hard pruning strategy of \citet{zhang2018graph} in their GCN model. They considered the full dependency tree to build the adjacency matrix but using a multi-head self attention-based soft pruning strategy, they can identify the important and unimportant edges in the graph. \newcite{Mandya2020GraphCO} proposed GCN over multiple sub-graphs for this task. They created such sub-graphs based on the shortest dependency path between two entities and the tokens associated with the two entities.

\begin{figure}[ht]
\centering
\includegraphics[scale=0.18]{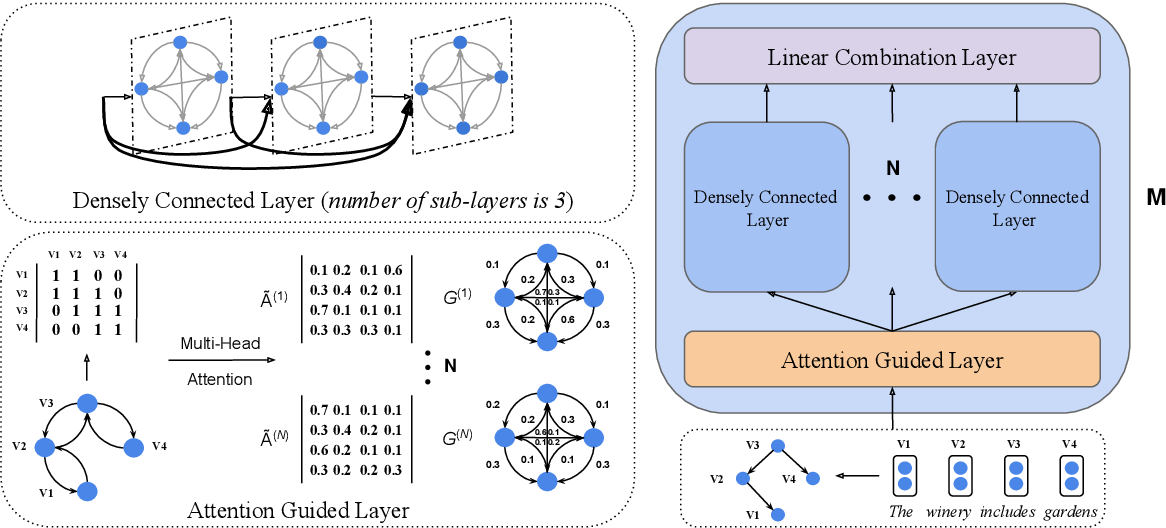}
\caption{The architecture of the attention guided graph convolutional network for relation extraction \cite{guo2019aggcn}.}
\label{fig:gcn}
\end{figure}

\newcite{sahu2019inter,christopoulou2019connecting,Nan2020ReasoningWL} used GCN for document-level relation extraction. \newcite{sahu2019inter} considered each token in a document as a node in a graph. They used syntactic dependency tree edges, word adjacency edges, and coreference edges to create the connections among the nodes. \newcite{christopoulou2019connecting} considered the entity mentions, entities, and sentences in a document as nodes of a graph. They used rule-based heuristics to create the edges among these nodes. In their graph, each node and each edge were represented by vectors. GCN was used to update the vectors of nodes and edges. Finally, the edge vector between the two concerned entities was passed to a classifier to find the relation. \newcite{Nan2020ReasoningWL} considered the entity mentions, entities, and tokens on the shortest dependency path between entity mentions as nodes in a graph. They used a structure induction module to learn the latent structure of the document-level graph. A multi-hop reasoning module was used to perform inference on the induced latent structure, where representations of the nodes were updated based on an information aggregation scheme. \newcite{Zeng2020DoubleGB} proposed a graph aggregation and inference network for document-level relation extraction. They construct an entity-mention level graph to capture their interaction in the document and an entity-level graph to aggregate the mention-level information. \newcite{Wang2020GlobaltoLocalNN} used a global graph similar to \newcite{christopoulou2019connecting} to model the entities in a document and then used a multi-head attention network \cite{vaswani2017attention} to aggregate the information in the global graph. \newcite{Zhou2020GlobalCG} proposed multi-head attention guided graph convolution network and \newcite{Li2020GraphED} proposed GCN-based dual attention network for document level relation extraction.

\subsubsection{Contextualized Embedding-Based Neural Models}

Contextualized word embeddings such as ELMo \citep{peters2018deep}, BERT \citep{devlin2019bert}, and SpanBERT \citep{Joshi2019SpanBERT} can be useful for relation extraction. These language models are trained on large corpora and can capture the contextual meaning of words in their vector representations. All neural models that are proposed for relation extraction use word representations such as Word2Vec \citep{mikolov2013distributed} or GloVe \citep{pennington2014glove} in their word embedding layer. Contextualized embeddings can be added in the embedding layer of the relation extraction models to improve their performance further. The SpanBERT model shows significant improvement in performance on the TACRED dataset. \citet{Joshi2019SpanBERT} replaced the entity 1 token with its type and SUBJ such as PER-SUBJ and entity 2 token with its type and OBJ such as LOC-OBJ in the sentences to train the model. Finally, they used a linear classifier on top of the CLS token vector to find the relation. \newcite{soares2019matching} also proposed a BERT based model where they used special marker for entity 1 and entity 2 in the sentences. Then they used the vector of the start token of the entity 1 and entity 2 for relation classification. 

\citet{Wang2019FinetuneBF} proposed two-step fine-tuning of BERT for document-level relation extraction on the DocRED dataset. In the first step, they used BERT to identify whether or not there is a relation between two entities. In the second step, they used BERT to classify the relation. \citet{Nan2020ReasoningWL} also used BERT in their model to show that it significantly improved the performance on the DocRED dataset compared to GloVe vectors. \citet{Han2020AND} used BERT to identify all possible relations among the entity pairs in documents in a single pass. They used entity types and special tokens to mark all the entity mentions in documents. All entity mentions of an entity received the same special token. Documents were passed to a pre-trained BERT model. An entity mention vector was obtained by averaging the BERT outputs of the entity mention tokens. An entity vector was obtained by averaging all the entity mention vectors of that entity. A bilinear classifier was used to classify the relation between two entities. \citet{Tang2020HINHI} proposed a hierarchical inference network for document-level relation extraction. They also showed that using BERT in their model improved performance significantly. 

\begin{table*}[ht]
\small
\centering
\begin{tabular}{l|ccc}
\hline
Model     & Prec. & Rec.  & F1   \\ \hline
SDP-LSTM \cite{xu2015classifying}  & 66.3 & 52.7 & 58.7 \\ 
Tree-LSTM \cite{tai2015improved} & 66.0 & 59.2 & 62.4 \\ 
GCN \cite{zhang2018graph}       & 69.8 & 59.0 & 64.0 \\ 
PA-LSTM \cite{zhang2017position}   & 65.7 & 64.5 & 65.1 \\ 
AGGCN \cite{guo2019aggcn}     & 69.9 & 60.9 & 65.1 \\ 
C-GCN \cite{zhang2018graph}     & 69.9 & 63.3 & 66.4 \\ 
GCN + PA-LSTM \cite{zhang2018graph}       & 71.7 & 63.0 & 67.1 \\ 
C-GCN + PA-LSTM \cite{zhang2018graph}     & 71.3 & 65.4 & 68.2 \\ 
C-AGGCN \cite{guo2019aggcn}  & 73.1 & 64.2 & 69.0 \\ \hline
BERT \cite{devlin2019bert}   & 69.1 & 63.9 & 66.4 \\ 
BERT$_{EM}$ \cite{soares2019matching} &   &  & 70.1 \\ 
SpanBERT \cite{Joshi2019SpanBERT} & 70.8 & 70.9 & 70.8 \\ 
BERT$_{EM}$ + MTB \cite{soares2019matching} &  &  & 71.5 \\ \hline
\end{tabular}
\caption{Current State-of-the-art on TACRED dataset.}
\label{tab:tacred_sota}
\end{table*}

\begin{table}[ht]
\small
\centering
\begin{tabular}{l|c}
\hline
Model     & F1   \\ \hline
SVM \cite{rink2010utd}  & 82.2 \\ 
CNN \cite{zeng2014relation}  & 82.7 \\ 
PA-LSTM \cite{zhang2017position}   & 82.7 \\ 
SDP-LSTM \cite{xu2015classifying}  & 83.7 \\
SPTree \cite{miwa2016end} & 84.4 \\ 
C-GCN \cite{zhang2018graph}     & 84.8 \\ 
C-AGGCN \cite{guo2019aggcn}  & 85.7 \\ 
Att-Input-CNN \cite{wang2016relation} & 87.5 \\ 
Att-Pooling-CNN \cite{wang2016relation} & 88.0 \\ \hline
BERT$_{EM}$ \cite{soares2019matching} & 89.2 \\ 
BERT$_{EM}$ + MTB \cite{soares2019matching} & 89.5 \\ \hline
\end{tabular}
\caption{Current State-of-the-art on SemEval 2010 Task 8 dataset.}
\label{tab:semeval_sota}
\end{table}

\begin{table*}[ht]
\small
\centering
\begin{tabular}{l|cccc}
\hline
Model   & Prec.     & Rec.      & F1   & Entity Matching Type     \\ \hline
Tagging \cite{zheng2017joint} & 0.624     & 0.317     & 0.420  & P     \\
CopyR \cite{zeng2018copyre}  & 0.610     & 0.566     & 0.587 & P     \\
GraphR \cite{fu2019graphrel}  & 0.639 & 0.600 & 0.619 & P  \\
CopyMTL$_{Mul}$ \cite{Zeng2020CopyMTLCM} & 0.757 & 0.687 & 0.720 & E \\
MrMep \cite{Chen2019MrMepJE}       & 0.779  & 0.766   & 0.771  & E    \\
HRL \cite{takanobu2019hrlre}       & 0.781    & 0.771    & 0.776  & E   \\
ETLSpan \cite{Bowen2020JointEO} & 0.855    & 0.717    & 0.780  & E   \\
PNDec \cite{nayak2019ptrnetdecoding}       & 0.806 & 0.773 & 0.789     & E    \\
WDec \cite{nayak2019ptrnetdecoding}       & 0.881 & 0.761 & 0.817     & E     \\
CasRel$_{LSTM}$ \cite{Wei2020ANC}        & 0.842    & 0.830   &  0.836    & P  \\
TPLinker$_{LSTM} $\cite{wang2020tplinker} & 0.860 & 0.820 & 0.840 & E  \\ 
RSAN \cite{Yuan2020ARA} & 0.857    & 0.836    & 0.846  & E   \\ 
RIN \cite{Sun2020RecurrentIN} & 0.839 & 0.855 & 0.847 & E \\ \hline
CGT$_{BERT}$ \cite{ye2020contrastive}  & 0.947 & 0.842 & 0.891 & E \\
CasRel$_{BERT}$ \cite{Wei2020ANC}        & 0.897    & 0.895   &  0.896    & P   \\ 
TPLinker$_{BERT} $\cite{wang2020tplinker} & 0.914 & 0.926 & 0.920 & E  \\ 
SPN$_{BERT}$ \cite{sui2020jointea} & 0.925 & 0.922 & 0.923 & E  \\ \hline
\end{tabular}
\caption{Current state-of-the-art performance on NYT24 datasets for the joint extraction task. P=Partial entity matching, E=Exact entity matching.}
\label{tab:nyt24_joint_sota}
\end{table*}

\begin{table*}[ht]
\small
\centering
\begin{tabular}{l|cccc}
\hline
Model    & Prec.      & Rec.      & F1   & Entity Matching Type  \\ \hline
Tagging \cite{zheng2017joint}     & 0.593     & 0.381     & 0.464     & E        \\
CopyR \cite{zeng2018copyre}  &  0.569    &  0.452    &  0.504     &   P          \\
SPTree \cite{miwa2016end}   & 0.492 & 0.557 & 0.522   & E     \\
HRL  \cite{takanobu2019hrlre}   & 0.692 & 0.601 & 0.643     & E   \\
MrMep \cite{Chen2019MrMepJE}    & 0.717    & 0.635  & 0.673   & E    \\
PNDec \cite{nayak2019ptrnetdecoding} & 0.732 & 0.624 & 0.673  & E \\ 
WDec \cite{nayak2019ptrnetdecoding} & 0.777 & 0.608 & 0.682     & E   \\ \hline
\end{tabular}
\caption{Current state-of-the-art performance on NYT29 datasets for the joint extraction task.}
\label{tab:nyt29_joint_sota}
\end{table*}

\begin{table*}[ht]
\small
\centering
\begin{tabular}{l|cccc}
\hline
Model   & Prec.     & Rec.      & F1     & Entity Matching Type     \\ \hline
Tagging \cite{zheng2017joint} & 0.525     & 0.193     & 0.283   & P    \\
CopyR \cite{zeng2018copyre}  & 0.377     & 0.364     & 0.371    & P  \\
GraphR \cite{fu2019graphrel}  & 0.447 & 0.411 & 0.429  & P  \\
CopyMTL$_{One}$ \cite{Zeng2020CopyMTLCM} & 0.578 & 0.601 & 0.589 & E \\
HRL \cite{takanobu2019hrlre}   & 0.695   & 0.629    &  0.660   & E   \\
MrMep \cite{Chen2019MrMepJE}   & 0.694    & 0.770  &  0.730  & E    \\
RIN \cite{Sun2020RecurrentIN} & 0.773 & 0.768 & 0.770 & E \\
RSAN \cite{Yuan2020ARA} & 0.805    & 0.838    & 0.821  & E \\
ETLSpan \cite{Bowen2020JointEO}  & 0.843    & 0.820    & 0.831  & E \\
CasRel$_{LSTM}$ \cite{Wei2020ANC}        & 0.869    & 0.806   &  0.837  & P \\ 
TPLinker$_{LSTM} $\cite{wang2020tplinker} & 0.919 & 0.816 & 0.864 & E \\ \hline
CGT$_{BERT}$ \cite{ye2020contrastive}  & 0.929 & 0.756 & 0.834 & E \\
TPLinker$_{BERT} $\cite{wang2020tplinker} & 0.889 & 0.845 & 0.867 & E \\ 
CasRel$_{BERT}$ \cite{Wei2020ANC}        & 0.934    & 0.901   &  0.918  & P \\ 
SPN$_{BERT}$ \cite{sui2020jointea} & 0.931 & 0.936 & 0.934 & P  \\\hline
\end{tabular}
\caption{Current state-of-the-art performance on WebNLG datasets for the joint extraction task.}
\label{tab:webnlg_joint_sota}
\end{table*}

\subsection{Noise Mitigation for Distantly Supervised Data}

The presence of noisy samples in distantly supervised data adversely affects the performance of models. Researchers have used different techniques in their models to mitigate the effects of noisy samples to make them more robust. Multi-instance relation extraction is one of the popular methods for noise mitigation. \citet{riedel2010modeling}, \citet{hoffmann2011knowledge}, \citet{mimlre}, \citet{lin2016neural}, \citet{yaghoobzadeh2017noise}, \citet{vashishth2018reside}, \citet{wu2018improving}, and \citet{ye2019distant} used this multi-instance learning concept in their proposed relation extraction models. For each entity pair, they used all the sentences that contained these two entities to find the relation between them. Their goal was to reduce the effect of noisy samples using this multi-instance setting. They used different types of sentence selection mechanisms to give importance to the sentences that contained relation-specific keywords and ignored the noisy sentences. \citet{ren2017cotype} and \citet{yaghoobzadeh2017noise} used the multi-task learning approach for mitigating the influence of noisy samples. They used fine-grained entity typing as an extra task in their model.

\begin{figure}[ht]
\centering
\includegraphics[scale=0.4]{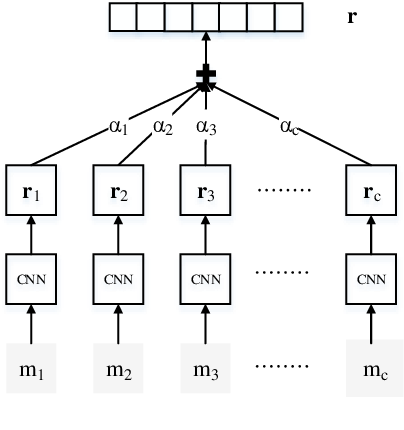}
\caption{The architecture of the attention over sentences model for bag-level relation extraction \cite{lin2016neural}.}
\label{fig:bag_att}
\end{figure}

\citet{wu2017adversarial} used an adversarial training approach for the same purpose. They added noise to the word embeddings to make the model more robust for distantly supervised training. \citet{Qin2018DSGANGA} used a generative adversarial network (GAN) to address the issue of noisy samples in relation extraction. They used a separate binary classifier as a generator in their model for each positive relation class to identify the true positives for that relation and filter out the noisy ones. \citet{qin2018robust} used reinforcement learning to identify the noisy samples for the positive relation classes. \newcite{jia2019arnor} proposed an attention-based regularization mechanism to address the noisy samples issue in distantly supervised relation extraction. They used the attention to identify the relation patterns in the sentences and sentences which do not contain such patterns are considered as noisy samples. \citet{He2020ImprovingNR} used reinforcement learning to identify the noisy samples for the positive relations and then used the identified noisy samples as unlabeled data in their model. \citet{Shang2020AreNS} used a clustering approach to identify the noisy samples. They assigned the correct relation label to these noisy samples and used them as additional training data in their model. 

\subsection{Zero-Shot and Few-Shot Relation Extraction}

Distantly supervised datasets cover a small subset of relations from the KBs. Existing KBs such as Freebase, Wikidata, and DBpedia contain thousands of relations. Due to the mismatch of the surface form of entities in KBs and texts, distant supervision cannot find adequate training samples for most relations in KBs. It means that distantly supervised models cannot fill the missing links belonging to these uncovered relations. Zero-shot or few-shot relation extraction can address this problem. These models can be trained on a set of relations and can be used for inferring another set of relations. 

\citet{Levy2017ZeroShotRE} and \citet{Li2019EntityRelationEA} converted the relation extraction task to a question-answering task and used the reading comprehension approach for zero-shot relation extraction. In this approach, entity 1 and the relation are used as questions, and entity 2 is the answer to the question. If entity 2 does not exist, the answer is {\em NIL}. \citet{Levy2017ZeroShotRE} used the BiDAF model \citep{seo2016bidirectional} with an additional {\em NIL} node in the output layer for this task on the WikiReading \citep{hewlett2016wikireading} dataset with additional negative samples. They used a set of relations during training and another set of relations during testing. \citet{Li2019EntityRelationEA} used templates to create the question using entity 1 and the relation. They modified the machine-reading comprehension models to a sequence tagging model so that they can find multiple answers to a question. Although they did not experiment with the zero-shot scenario, this approach can be used for zero-shot relation extraction too. FewRel 2.0 \citep{gao2019fewrel} is a dataset for few-shot relation extraction. In few-shot relation extraction, training and test relations are different just like zero-shot extraction. But during testing, a few examples of the test relations are provided to the model for better prediction. 

\subsection{Joint Extraction Approaches}

All the previously mentioned works on relation extraction assume that entities are already identified by a named entity recognition system. They classify the relation between two given entities at the sentence level or the bag-of-sentences level. These models depend on an external named entity recognition system to identify the entities in a text. Recently, some researchers \citep{katiyar2016investigating,miwa2016end,bekoulis2018joint,dat2019end} tried to remove this dependency. They tried to bring the entity recognition and relation identification tasks closer by sharing their parameters and optimizing them together. They first identify all the entities in a sentence and then find the relation among all the pairs of identified entities. Although they identify the entities and relations in the same network, they still identify the entities first and then determine the relation among all possible pairs in the same network. So these models miss the interaction among the relation triplets present in a sentence. These approaches resemble the pipeline approach to some extent.

\citet{zheng2017joint} first proposed a truly joint extraction model for this task. They used a sequence tagging scheme to jointly extract the entities and relations. They created a set of tags derived from the Cartesian product of entity tags and relation tags. These new tags can encode the entity information and relation information together. But this strategy does not work when entities are shared among multiple triplets, as only one tag can be assigned to a token. \citet{zeng2018copyre} proposed an encoder-decoder model with a copy mechanism to extract relation triplets with overlapping entities. Their model has a copy network to copy the last token of two entities from the source sentence and a classification network to classify the relation between copied tokens. Their model cannot extract the full entity names of the triplets. Their best performing model uses a separate decoder to extract each triplet. During training, they need to fix the maximum number of decoders and during inference, their model can only extract up to that fixed number of triplets. Also, due to the use of separate decoders for each triplet, their model misses the interaction among the triplets. 

\begin{figure}[ht]
\centering
\includegraphics[scale=0.3]{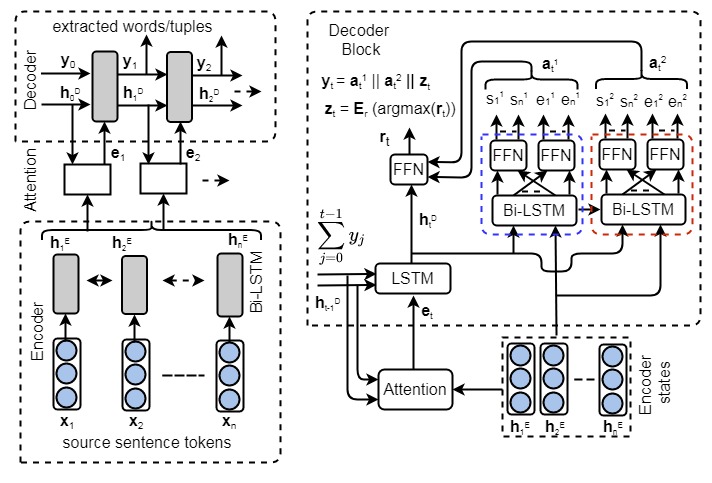}
\caption{The architecture of the joint entity and relation extraction model as proposed in \newcite{nayak2019ptrnetdecoding}.}
\label{fig:ptrnet}
\end{figure}

\begin{figure*}[ht]
\centering
\includegraphics[scale=0.45]{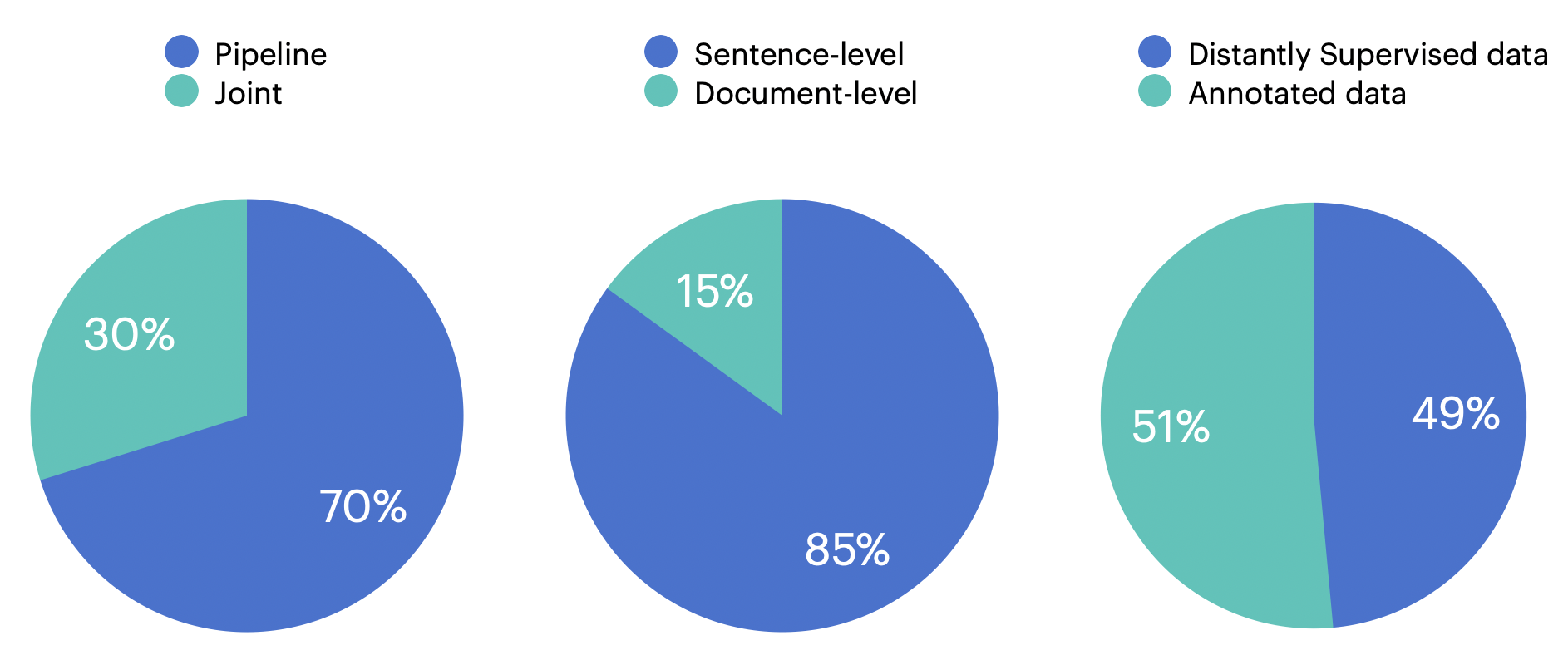}
\caption{The statistics of the research articles published in year 2019 (CoNLL, ACL, EMNLP, AAAI, IJCAI) and 2020 (COLING, ACL, EMNLP, AAAI, IJCAI). The left one shows the pipeline vs joint extraction models, the middle one shows the sentence-level vs document-level extraction models, and the right one shows the use of distantly supervised datasets vs annotated datasets.}
\label{fig:chart}
\end{figure*}

\citet{takanobu2019hrlre} proposed a hierarchical reinforcement learning-based (RL) deep neural model for joint entity and relation extraction. A high-level RL is used to identify the relation based on the relation-specific tokens in the sentences. After a relation is identified, a low-level RL is used to extract the two entities associated with the relation using a sequence labeling approach. This process is repeated multiple times to extract all the relation triplets present in the sentences. A special {\em None} relation is used to identify no relation situation in the sentences. Entities extracted associated with the {\em None} relations are ignored. \citet{fu2019graphrel} used a graph convolutional network (GCN) where they treated each token in a sentence as a node in a graph and edges were considered as relations. \citet{trisedya2019neural} used an N-gram attention mechanism with an encoder-decoder model for the completion of knowledge bases using distantly supervised data. \citet{Chen2019MrMepJE} used the encoder-decoder framework for this task where they used a CNN-based multi-label classifier to find all the relations first, then used multi-head attention \citep{vaswani2017attention} to extract the entities corresponding to each relation. \newcite{nayak2019ptrnetdecoding} used encoder-decoder network for this joint extraction task. They proposed a word-level decoding framework and a pointer network-based decoding framework for the same. 

CopyMTL model \cite{Zeng2020CopyMTLCM} was proposed to address the issues of CopyR \citep{zeng2018copyre} model. CopyR model can only extract the last token of the entities, whereas CopyMTL model used a sequence tagging approach to extract the full entity names. \citet{Bowen2020JointEO} decomposed the joint extraction task into two sub-tasks: (i) head entity extraction (ii) tail entity and relation extraction. They used a sequence tagging approach to solve these two sub-tasks. Similarly, \citet{Wei2020ANC} proposed a sequence tagging approach for this task. They first identified the head entities and then for each head entity and each relation, they identified the tail entities using a sequence tagging approach. They used pre-trained BERT \citep{devlin2019bert} in their model to improve the performance. \newcite{Yuan2020ARA} used a relation-specific attention mechanism with sequence labeling to jointly extract the entities and relations. \newcite{wang2020tplinker} proposed a single-stage joint extraction model using entity-pair linking. They aligned the sentence tokens using the Cartesian product so that the boundary tokens of the subject and object entities are aligned. Then they used a classifier to tag each token-pair as entity head, entity tail, subject head, subject tail, object head, and object tail for each relation separately. This scheme can identity the multiple triplets with overlapping entities easily. \newcite{sui2020jointea} proposed a bipartite matching loss in the encoder-decoder network which considers the group of relation triplets as a set, not as a sequence. \newcite{ye2020contrastive} transformer-based generative model for this task. They used negative triplets to train the transformer model in contrastive settings. \newcite{Wang2020TwoAB} proposed a table-sequence encoder model where the sequence encoder captures the entity-related information and the table encoder captures the relation-specific information. \newcite{Sun2020RecurrentIN} proposed a recurrent multi-task learning architecture to explicitly capture the interaction between entity recognition task and relation classification task. \newcite{Ji2020SpanbasedJE} proposed a span-based multi-head attention network for joint extraction task. Each text span is a candidate entity and each text span pairs is a candidate for relation triplets.

\section{Current State-of-the-art \& Trends}

NYT10 is the most popular dataset for experiments in pipeline-based relation extraction. Since the test dataset of NYT10 is not manually annotated, researchers mostly report a precision-recall curve to compare the models \cite{vashishth2018reside,ye2019distant,Li2020SelfAttentionES}. TACRED and SemEval 2010 Task 8 datasets are manually annotated and can be used for automatic evaluation. We have included the current state-of-the-art on these two dataset in Table \ref{tab:tacred_sota} and Table \ref{tab:semeval_sota}. DocRED\footnote{https://competitions.codalab.org/competitions/20717} and FewRel\footnote{https://thunlp.github.io/2/fewrel2\_nota.html} datasets have manually annotated testset and their they have a leaderboard where current state-of-the-art can be found. For the joint extraction task researchers used NYT24, NYT29, and WebNLG datasets which have a considerably large number of relations. We have included the current state-of-the-art performance of the models on NYT24, NYT29, and WebNLG datasets in Table \ref{tab:nyt24_joint_sota}, Table \ref{tab:nyt29_joint_sota}, and Table \ref{tab:webnlg_joint_sota} respectively.
\begin{figure}[ht!]
\centering
\includegraphics[scale=0.5]{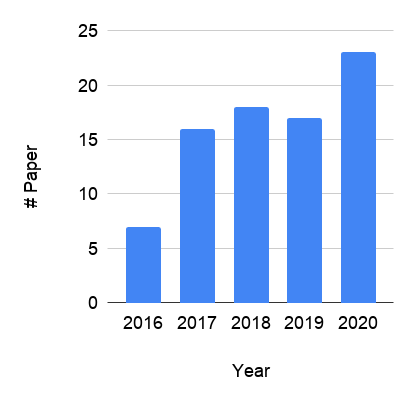}
\caption{Publication trend of relation extraction research at ACL, EMNLP, AAAI, and IJCAI in 2016--2020.}
\label{fig:5year}
\end{figure}


We analyze the research articles published in 2019 (CoNLL, ACL, EMNLP, AAAI, IJCAI) and 2020 (COLING, ACL, EMNLP, AAAI, IJCAI) and include statistics in Figure \ref{fig:chart}. We see that majority of the research focuses on pipeline-based approaches on sentence-level relation extraction. We also see that the use of distantly supervised datasets and annotated datasets for experiments is evenly distributed among the published articles. We also show the increasing trends of yearly publications in relation extraction in Figure \ref{fig:5year} over the last 5 years period (2016-2020).

\section{Future Research Directions}

With the progress of deep learning algorithms, significant advances have been made in the relation extraction task. However, many challenges remain in this area. In the pipeline approaches, since we need to find relations among all pairs of entities, there can be a very large number of {\em None} instances. This {\em None} class is challenging to identify as it is not a single relation but any relation outside the set of positive relations. Erroneous detection of {\em None} relation reduces the precision of the model and can add many wrong triplets to the KB. To build a cleaner KB, models have to perform very well to detect the {\em None} relation along with classifying the positive relations correctly. 

Regarding the joint extraction approach, researchers do not include sentences with zero triplets in training or testing. NYT24 and NYT29 datasets are created after removing the sentences with zero triplets from the original NYT11 and NYT10 datasets. NYT11 and NYT10 datasets contain many sentences that do not have any relation triplets. So in the future, detecting sentences with no relation triplets must be handled in the joint extraction approaches. 

Current relation extraction models deal with very few relations whereas existing knowledge bases have thousands of relations. In the future, we should focus more on document-level relation extraction or possibly relation extraction across documents. Following the idea proposed in \newcite{nayak2020deep}, we should extend the task of relation extraction to N-hop to cover more relations from the KB. However, it may not be easy to extend the task as the inclusion of more documents in the chain may make the data noisier. It will be challenging to create a clean dataset for N-hop relation extraction. Also, we need to explore zero-shot or few-shot relation extraction to cover the relations for which we cannot obtain enough training data using distant supervision. 

\section{Conclusion}

In this survey paper, we detail the recent progress in neural network-based relation extraction research that includes both pipeline-based and joint extraction-based relation extraction approaches. Furthermore, we describe different relation-extraction datasets and setup baselines to facilitate future research. Key issues with the current distantly-supervised datasets are also pointed out. We finally conclude with the possible future research directions to advance this field.

\bibliographystyle{acl_natbib}
\bibliography{anthology,acl2021}

\begin{thebibliography}{127}
\expandafter\ifx\csname natexlab\endcsname\relax\def\natexlab#1{#1}\fi

\bibitem[{Andrew(2018)}]{andrew2018automatic}
Judith~Jeyafreeda Andrew. 2018.
\newblock Automatic extraction of entities and relation from legal documents.
\newblock In \emph{Proceedings of the Seventh Named Entities Workshop}.

\bibitem[{Baldini~Soares et~al.(2019)Baldini~Soares, FitzGerald, Ling, and
  Kwiatkowski}]{soares2019matching}
Livio Baldini~Soares, Nicholas FitzGerald, Jeffrey Ling, and Tom Kwiatkowski.
  2019.
\newblock Matching the blanks: Distributional similarity for relation learning.
\newblock In \emph{ACL}.

\bibitem[{Banko et~al.(2007)Banko, Cafarella, Soderland, Broadhead, and
  Etzioni}]{banko2007open}
Michele Banko, Michael~J Cafarella, Stephen Soderland, Matthew Broadhead, and
  Oren Etzioni. 2007.
\newblock Open information extraction from the web.
\newblock In \emph{IJCAI}.

\bibitem[{Bekoulis et~al.(2018)Bekoulis, Deleu, Demeester, and
  Develder}]{bekoulis2018joint}
Giannis Bekoulis, Johannes Deleu, Thomas Demeester, and Chris Develder. 2018.
\newblock Joint entity recognition and relation extraction as a multi-head
  selection problem.
\newblock \emph{Expert Systems with Applications}.

\bibitem[{Bizer et~al.(2009)Bizer, Lehmann, Kobilarov, Auer, Becker, Cyganiak,
  and Hellmann}]{bizer2009dbpedia}
Christian Bizer, Jens Lehmann, Georgi Kobilarov, S{\"o}ren Auer, Christian
  Becker, Richard Cyganiak, and Sebastian Hellmann. 2009.
\newblock {DB}pedia-{A} crystallization point for the web of data.
\newblock \emph{Web Semantics: Science, Services and Agents on the World Wide
  Web}.

\bibitem[{Bollacker et~al.(2008)Bollacker, Evans, Paritosh, Sturge, and
  Taylor}]{bollacker2008freebase}
Kurt Bollacker, Colin Evans, Praveen Paritosh, Tim Sturge, and Jamie Taylor.
  2008.
\newblock Freebase: {A} collaboratively created graph database for structuring
  human knowledge.
\newblock In \emph{SIGMOD}.

\bibitem[{Bowen et~al.(2019)Bowen, Zhang, Liu, Wang, Li, and
  Li}]{Bowen2019BeyondWA}
Yu~Bowen, Zhenyu Zhang, Tingwen Liu, Bin Wang, Sujian Li, and Q.~Li. 2019.
\newblock Beyond word attention: Using segment attention in neural relation
  extraction.
\newblock In \emph{IJCAI}.

\bibitem[{Bowen et~al.(2020)Bowen, Zhang, Su, Wang, Liu, Wang, and
  Li}]{Bowen2020JointEO}
Yu~Bowen, Zhenyu Zhang, Jianlin Su, Yubin Wang, Tingwen Liu, Bin Wang, and
  Sujian Li. 2020.
\newblock Joint extraction of entities and relations based on a novel
  decomposition strategy.
\newblock In \emph{ECAI}.

\bibitem[{Chen et~al.(2019)Chen, Yuan, Wang, and Bai}]{Chen2019MrMepJE}
Jiayu Chen, Caixia Yuan, Xiao-Jie Wang, and Ziwei Bai. 2019.
\newblock {MrMep}: {J}oint extraction of multiple relations and multiple entity
  pairs based on triplet attention.
\newblock In \emph{CoNLL}.

\bibitem[{Chiu and Nichols(2016)}]{chiu2016named}
Jason Chiu and Eric Nichols. 2016.
\newblock Named entity recognition with bidirectional {LSTM-CNN}s.
\newblock In \emph{TACL}.

\bibitem[{Cho et~al.(2014)Cho, Van~Merri{\"e}nboer, Bahdanau, and
  Bengio}]{cho2014properties}
Kyunghyun Cho, Bart Van~Merri{\"e}nboer, Dzmitry Bahdanau, and Yoshua Bengio.
  2014.
\newblock On the properties of neural machine translation: Encoder-decoder
  approaches.
\newblock In \emph{Workshop on Syntax, Semantics and Structure in Statistical
  Translation}.

\bibitem[{Choi(2018)}]{Choi2018ExtractionOP}
Sung-Pil Choi. 2018.
\newblock Extraction of protein–protein interactions (ppis) from the
  literature by deep convolutional neural networks with various feature
  embeddings.
\newblock \emph{Journal of Information Science}.

\bibitem[{Christensen et~al.(2011)Christensen, Mausam, Soderland, and
  Etzioni}]{christensen2011srlie}
Janara Christensen, Mausam, Stephen Soderland, and Oren Etzioni. 2011.
\newblock An analysis of open information extraction based on semantic role
  labeling.
\newblock In \emph{K-CAP}.

\bibitem[{Christopoulou et~al.(2019)Christopoulou, Miwa, and
  Ananiadou}]{christopoulou2019connecting}
Fenia Christopoulou, Makoto Miwa, and Sophia Ananiadou. 2019.
\newblock Connecting the dots: Document-level neural relation extraction with
  edge-oriented graphs.
\newblock In \emph{EMNLP and IJCNLP}.

\bibitem[{Cui et~al.(2017)Cui, Li, Wang, and You}]{Cui2017ASO}
Meiji Cui, L.~Li, Zhihong Wang, and Mingyu You. 2017.
\newblock A survey on relation extraction.
\newblock In \emph{CCKS}.

\bibitem[{Devlin et~al.(2019)Devlin, Chang, Lee, and
  Toutanova}]{devlin2019bert}
Jacob Devlin, Ming-Wei Chang, Kenton Lee, and Kristina Toutanova. 2019.
\newblock {BERT}: Pre-training of deep bidirectional transformers for language
  understanding.
\newblock In \emph{NAACL-HLT}.

\bibitem[{Doddington et~al.(2004)Doddington, Mitchell, Przybocki, Ramshaw,
  Strassel, and Weischedel}]{doddington2004automatic}
George~R Doddington, Alexis Mitchell, Mark~A Przybocki, Lance~A Ramshaw,
  Stephanie~M Strassel, and Ralph~M Weischedel. 2004.
\newblock The automatic content extraction ({ACE}) program-tasks, data, and
  evaluation.
\newblock In \emph{LREC}.

\bibitem[{Etzioni et~al.(2004)Etzioni, Cafarella, Downey, Kok, Popescu, Shaked,
  Soderland, Weld, and Yates}]{etzioni2004knowitall}
Oren Etzioni, Michael Cafarella, Doug Downey, Stanley Kok, Ana-Maria Popescu,
  Tal Shaked, Stephen Soderland, Daniel~S Weld, and Alexander Yates. 2004.
\newblock Web-scale information extraction in {K}now{I}t{A}ll:(preliminary
  results).
\newblock In \emph{WWW}.

\bibitem[{Etzioni et~al.(2011)Etzioni, Fader, Christensen, Soderland, and
  Mausam}]{etzioni2011reverb}
Oren Etzioni, Anthony Fader, Janara Christensen, Stephen Soderland, and Mausam.
  2011.
\newblock Open information extraction: The second generation.
\newblock In \emph{IJCAI}.

\bibitem[{Fellbaum(2000)}]{Fellbaum2000WordNetA}
Christiane Fellbaum. 2000.
\newblock {WordNet}: {A}n electronic lexical database.
\newblock \emph{Language}.

\bibitem[{Fu et~al.(2019)Fu, Li, and Ma}]{fu2019graphrel}
Tsu-Jui Fu, Peng-Hsuan Li, and Wei-Yun Ma. 2019.
\newblock {G}raph{R}el: Modeling text as relational graphs for joint entity and
  relation extraction.
\newblock In \emph{ACL}.

\bibitem[{Gao et~al.(2019)Gao, Han, Zhu, Liu, Li, Sun, and
  Zhou}]{gao2019fewrel}
Tianyu Gao, Xu~Han, Hao Zhu, Zhiyuan Liu, Peng Li, Maosong Sun, and Jie Zhou.
  2019.
\newblock {F}ew{R}el 2.0: Towards more challenging few-shot relation
  classification.
\newblock In \emph{EMNLP and IJCNLP}.

\bibitem[{Gardent et~al.(2017)Gardent, Shimorina, Narayan, and
  Perez-Beltrachini}]{Gardent2017CreatingTC}
Claire Gardent, Anastasia Shimorina, Shashi Narayan, and Laura
  Perez-Beltrachini. 2017.
\newblock Creating training corpora for nlg micro-planners.
\newblock In \emph{ACL}.

\bibitem[{Gu et~al.(2016)Gu, Qian, and Zhou}]{Gu2016ChemicalinducedDR}
Jinghang Gu, Longhua Qian, and Guodong Zhou. 2016.
\newblock Chemical-induced disease relation extraction with various linguistic
  features.
\newblock \emph{Database: The Journal of Biological Databases and Curation}.

\bibitem[{Guo et~al.(2019)Guo, Zhang, and Lu}]{guo2019aggcn}
Zhijiang Guo, Yan Zhang, and Wei Lu. 2019.
\newblock Attention guided graph convolutional networks for relation
  extraction.
\newblock In \emph{ACL}.

\bibitem[{Han and Wang(2020)}]{Han2020AND}
Xiaoyu Han and Lei Wang. 2020.
\newblock A novel document-level relation extraction method based on {BERT} and
  entity information.
\newblock \emph{IEEE Access}.

\bibitem[{Han et~al.(2020)Han, Gao, Lin, Peng, Yang, Xiao, Liu, Li, Sun, and
  Zhou}]{Han2020MoreDM}
Xu~Han, Tianyu Gao, Yankai Lin, H.~Peng, Y.~Yang, Chaojun Xiao, Zhiyuan Liu,
  Peng Li, Maosong Sun, and Jie Zhou. 2020.
\newblock More data, more relations, more context and more openness: A review
  and outlook for relation extraction.
\newblock In \emph{AACL and IJCNLP}.

\bibitem[{He et~al.(2020)He, Chen, Wang, Zhang, Wang, and
  Zhang}]{He2020ImprovingNR}
Zhengqiu He, Wenliang Chen, Yuyi Wang, Wei Zhang, Guanchun Wang, and Min Zhang.
  2020.
\newblock Improving neural relation extraction with positive and unlabeled
  learning.
\newblock In \emph{AAAI}.

\bibitem[{Hendrickx et~al.(2010)Hendrickx, Kim, Kozareva, Nakov,
  {\'O}~S{\'e}aghdha, Pad{\'o}, Pennacchiotti, Romano, and
  Szpakowicz}]{hendrickx2010semeval}
Iris Hendrickx, Su~Nam Kim, Zornitsa Kozareva, Preslav Nakov, Diarmuid
  {\'O}~S{\'e}aghdha, Sebastian Pad{\'o}, Marco Pennacchiotti, Lorenza Romano,
  and Stan Szpakowicz. 2010.
\newblock Semeval-2010 task 8: Multi-way classification of semantic relations
  between pairs of nominals.
\newblock In \emph{SemEval}.

\bibitem[{Hewlett et~al.(2016)Hewlett, Lacoste, Jones, Polosukhin, Fandrianto,
  Han, Kelcey, and Berthelot}]{hewlett2016wikireading}
Daniel Hewlett, Alexandre Lacoste, Llion Jones, Illia Polosukhin, Andrew
  Fandrianto, Jay Han, Matthew Kelcey, and David Berthelot. 2016.
\newblock {W}iki{R}eading: {A} novel large-scale language understanding task
  over {W}ikipedia.
\newblock In \emph{ACL}.

\bibitem[{Hochreiter and Schmidhuber(1997)}]{hochreiter1997long}
Sepp Hochreiter and J{\"u}rgen Schmidhuber. 1997.
\newblock Long short-term memory.
\newblock \emph{Neural Computation}.

\bibitem[{Hoffmann et~al.(2011)Hoffmann, Zhang, Ling, Zettlemoyer, and
  Weld}]{hoffmann2011knowledge}
Raphael Hoffmann, Congle Zhang, Xiao Ling, Luke Zettlemoyer, and Daniel~S Weld.
  2011.
\newblock Knowledge-based weak supervision for information extraction of
  overlapping relations.
\newblock In \emph{ACL}.

\bibitem[{huang et~al.(2020)huang, Wu, and Wang}]{huang2020KnowledgeGA}
Luyang huang, L.~Wu, and L.~Wang. 2020.
\newblock Knowledge graph-augmented abstractive summarization with
  semantic-driven cloze reward.
\newblock In \emph{ACL}.

\bibitem[{Huang et~al.(2015)Huang, Xu, and Yu}]{Huang2015BidirectionalLM}
Zhiheng Huang, Wei Xu, and Kai Yu. 2015.
\newblock Bidirectional {LSTM-CRF} models for sequence tagging.
\newblock \emph{ArXiv}.

\bibitem[{Jain et~al.(2020)Jain, van Zuylen, Hajishirzi, and
  Beltagy}]{jain2020scirex}
Sarthak Jain, Madeleine van Zuylen, Hannaneh Hajishirzi, and Iz~Beltagy. 2020.
\newblock {S}ci{REX}: {A} challenge dataset for document-level information
  extraction.
\newblock In \emph{ACL}.

\bibitem[{Jat et~al.(2017)Jat, Khandelwal, and Talukdar}]{jat2018attention}
Sharmistha Jat, Siddhesh Khandelwal, and Partha Talukdar. 2017.
\newblock Improving distantly supervised relation extraction using word and
  entity based attention.
\newblock In \emph{AKBC}.

\bibitem[{Ji et~al.(2020)Ji, Yu, Li, Ma, Wu, Tan, and Liu}]{Ji2020SpanbasedJE}
Bin Ji, Jie Yu, Shasha Li, Jun Ma, Q.~Wu, Yusong Tan, and Huijun Liu. 2020.
\newblock Span-based joint entity and relation extraction with attention-based
  span-specific and contextual semantic representations.
\newblock In \emph{COLING}.

\bibitem[{Jia et~al.(2019)Jia, Dai, Xiao, and Wu}]{jia2019arnor}
Wei Jia, Dai Dai, Xinyan Xiao, and Hua Wu. 2019.
\newblock {ARNOR}: Attention regularization based noise reduction for distant
  supervision relation classification.
\newblock In \emph{ACL}.

\bibitem[{Joshi et~al.(2019)Joshi, Chen, Liu, Weld, Zettlemoyer, and
  Levy}]{Joshi2019SpanBERT}
Mandar Joshi, Danqi Chen, Yinhan Liu, Daniel~S. Weld, Luke Zettlemoyer, and
  Omer Levy. 2019.
\newblock {SpanBERT}: {I}mproving pre-training by representing and predicting
  spans.
\newblock \emph{TACL}.

\bibitem[{Katiyar and Cardie(2016)}]{katiyar2016investigating}
Arzoo Katiyar and Claire Cardie. 2016.
\newblock Investigating {LSTM}s for joint extraction of opinion entities and
  relations.
\newblock In \emph{ACL}.

\bibitem[{Kipf and Welling(2017)}]{Kipf2017SemiSupervisedCW}
Thomas Kipf and Max Welling. 2017.
\newblock Semi-supervised classification with graph convolutional networks.
\newblock In \emph{ICLR}.

\bibitem[{Kumar(2017)}]{Kumar2017ASO}
Shantanu Kumar. 2017.
\newblock A survey of deep learning methods for relation extraction.
\newblock \emph{ArXiv}.

\bibitem[{Lample et~al.(2016)Lample, Ballesteros, Subramanian, Kawakami, and
  Dyer}]{lample2016neural}
Guillaume Lample, Miguel Ballesteros, Sandeep Subramanian, Kazuya Kawakami, and
  Chris Dyer. 2016.
\newblock Neural architectures for named entity recognition.
\newblock In \emph{NAACL-HLT}.

\bibitem[{Levy et~al.(2017)Levy, Seo, Choi, and
  Zettlemoyer}]{Levy2017ZeroShotRE}
Omer Levy, Minjoon Seo, Eunsol Choi, and Luke~S. Zettlemoyer. 2017.
\newblock Zero-shot relation extraction via reading comprehension.
\newblock In \emph{CoNLL}.

\bibitem[{Li et~al.(2020{\natexlab{a}})Li, Ye, Sheng, Xie, Xi, and
  Zhang}]{Li2020GraphED}
Bo~Li, Wei Ye, Zhonghao Sheng, Rui Xie, Xiangyu Xi, and Shikun Zhang.
  2020{\natexlab{a}}.
\newblock Graph enhanced dual attention network for document-level relation
  extraction.
\newblock In \emph{COLING}.

\bibitem[{Li et~al.(2017)Li, Zhang, Fu, and Ji}]{Li2017ANJ}
Fei Li, Meishan Zhang, G.~Fu, and D.~Ji. 2017.
\newblock A neural joint model for entity and relation extraction from
  biomedical text.
\newblock \emph{BMC Bioinformatics}.

\bibitem[{Li et~al.(2019)Li, Yin, Sun, Li, Yuan, Chai, Zhou, and
  Li}]{Li2019EntityRelationEA}
Xiaoya Li, Fan Yin, Zijun Sun, Xiayu Li, Arianna Yuan, Duo Chai, Mingxin Zhou,
  and Jiwei Li. 2019.
\newblock Entity-relation extraction as multi-turn question answering.
\newblock In \emph{ACL}.

\bibitem[{Li et~al.(2020{\natexlab{b}})Li, Long, Shen, Zhou, Yao, Huo, and
  Jiang}]{Li2020SelfAttentionES}
Yang Li, Guodong Long, Tao Shen, Tianyi Zhou, L.~Yao, Huan Huo, and Jing Jiang.
  2020{\natexlab{b}}.
\newblock Self-attention enhanced selective gate with entity-aware embedding
  for distantly supervised relation extraction.
\newblock In \emph{AAAI}.

\bibitem[{Lin et~al.(2016)Lin, Shen, Liu, Luan, and Sun}]{lin2016neural}
Yankai Lin, Shiqi Shen, Zhiyuan Liu, Huanbo Luan, and Maosong Sun. 2016.
\newblock Neural relation extraction with selective attention over instances.
\newblock In \emph{ACL}.

\bibitem[{Liu et~al.(2019)Liu, Ott, Goyal, Du, Joshi, Chen, Levy, Lewis,
  Zettlemoyer, and Stoyanov}]{Liu2019RoBERTaAR}
Y.~Liu, Myle Ott, Naman Goyal, Jingfei Du, Mandar Joshi, Danqi Chen, Omer Levy,
  M.~Lewis, Luke Zettlemoyer, and Veselin Stoyanov. 2019.
\newblock Roberta: A robustly optimized bert pretraining approach.
\newblock \emph{ArXiv}.

\bibitem[{Liu et~al.(2015)Liu, Wei, Li, Ji, Zhou, and Wang}]{liu2015dependency}
Yang Liu, Furu Wei, Sujian Li, Heng Ji, Ming Zhou, and Houfeng Wang. 2015.
\newblock A dependency-based neural network for relation classification.
\newblock In \emph{ACL and IJCNLP}.

\bibitem[{Luan et~al.(2017)Luan, Ostendorf, and
  Hajishirzi}]{luan2017scientific}
Yi~Luan, Mari Ostendorf, and Hannaneh Hajishirzi. 2017.
\newblock Scientific information extraction with semi-supervised neural
  tagging.
\newblock In \emph{Proceedings of the 2017 Conference on Empirical Methods in
  Natural Language Processing}.

\bibitem[{Ma and Hovy(2016)}]{Ma2016EndtoendSL}
Xuezhe Ma and Eduard~H. Hovy. 2016.
\newblock End-to-end sequence labeling via bi-directional {LSTM}-{CNN}s-{CRF}.
\newblock In \emph{ACL}.

\bibitem[{Mandya et~al.(2020)Mandya, Bollegala, and Coenen}]{Mandya2020GraphCO}
Angrosh Mandya, Danushka Bollegala, and F.~Coenen. 2020.
\newblock Graph convolution over multiple dependency sub-graphs for relation
  extraction.
\newblock In \emph{COLING}.

\bibitem[{Mausam et~al.(2012)Mausam, Schmitz, Soderland, Bart, and
  Etzioni}]{schmitz2012ollie}
Mausam, Michael Schmitz, Stephen Soderland, Robert Bart, and Oren Etzioni.
  2012.
\newblock Open language learning for information extraction.
\newblock In \emph{EMNLP-CoNLL}.

\bibitem[{Mikolov et~al.(2013)Mikolov, Sutskever, Chen, Corrado, and
  Dean}]{mikolov2013distributed}
Tomas Mikolov, Ilya Sutskever, Kai Chen, Greg~S Corrado, and Jeff Dean. 2013.
\newblock Distributed representations of words and phrases and their
  compositionality.
\newblock In \emph{NIPS}.

\bibitem[{Mintz et~al.(2009)Mintz, Bills, Snow, and
  Jurafsky}]{mintz2009distant}
Mike Mintz, Steven Bills, Rion Snow, and Dan Jurafsky. 2009.
\newblock Distant supervision for relation extraction without labeled data.
\newblock In \emph{ACL and IJCNLP}.

\bibitem[{Miwa and Bansal(2016)}]{miwa2016end}
Makoto Miwa and Mohit Bansal. 2016.
\newblock End-to-end relation extraction using {LSTM}s on sequences and tree
  structures.
\newblock In \emph{ACL}.

\bibitem[{Nan et~al.(2020)Nan, Guo, Sekulic, and Lu}]{Nan2020ReasoningWL}
Guoshun Nan, Zhijiang Guo, Ivan Sekulic, and Wei Lu. 2020.
\newblock Reasoning with latent structure refinement for document-level
  relation extraction.
\newblock In \emph{ACL}.

\bibitem[{Nayak(2020)}]{nayak2020deep}
Tapas Nayak. 2020.
\newblock Deep neural networks for relation extraction.
\newblock \emph{NUS Scholar Bank}.

\bibitem[{Nayak and Ng(2019)}]{nayak2019effective}
Tapas Nayak and Hwee~Tou Ng. 2019.
\newblock Effective attention modeling for neural relation extraction.
\newblock In \emph{CoNLL}.

\bibitem[{Nayak and Ng(2020)}]{nayak2019ptrnetdecoding}
Tapas Nayak and Hwee~Tou Ng. 2020.
\newblock Effective modeling of encoder-decoder architecture for joint entity
  and relation extraction.
\newblock In \emph{AAAI}.

\bibitem[{Nguyen and Verspoor(2019)}]{dat2019end}
Dat~Quoc Nguyen and Karin Verspoor. 2019.
\newblock End-to-end neural relation extraction using deep biaffine attention.
\newblock In \emph{ECIR}.

\bibitem[{Pawar et~al.(2017)Pawar, Palshikar, and
  Bhattacharyya}]{Pawar2017RelationE}
S.~Pawar, Girish~Keshav Palshikar, and P.~Bhattacharyya. 2017.
\newblock Relation extraction : A survey.
\newblock \emph{ArXiv}.

\bibitem[{Peng et~al.(2017)Peng, Poon, Quirk, Toutanova, and
  Yih}]{peng2017cross}
Nanyun Peng, Hoifung Poon, Chris Quirk, Kristina Toutanova, and Wen-tau Yih.
  2017.
\newblock Cross-sentence n-ary relation extraction with graph {LSTM}s.
\newblock \emph{TACL}.

\bibitem[{Pennington et~al.(2014)Pennington, Socher, and
  Manning}]{pennington2014glove}
Jeffrey Pennington, Richard Socher, and Christopher Manning. 2014.
\newblock {GloVe}: Global vectors for word representation.
\newblock In \emph{EMNLP}.

\bibitem[{Peters et~al.(2018)Peters, Neumann, Iyyer, Gardner, Clark, Lee, and
  Zettlemoyer}]{peters2018deep}
Matthew Peters, Mark Neumann, Mohit Iyyer, Matt Gardner, Christopher Clark,
  Kenton Lee, and Luke Zettlemoyer. 2018.
\newblock Deep contextualized word representations.
\newblock In \emph{NAACL-HLT}.

\bibitem[{Qin et~al.(2018{\natexlab{a}})Qin, Xu, and Wang}]{Qin2018DSGANGA}
Pengda Qin, Weiran Xu, and William~Yang Wang. 2018{\natexlab{a}}.
\newblock {DSGAN}: {G}enerative adversarial training for distant supervision
  relation extraction.
\newblock In \emph{ACL}.

\bibitem[{Qin et~al.(2018{\natexlab{b}})Qin, Xu, and Wang}]{qin2018robust}
Pengda Qin, Weiran Xu, and William~Yang Wang. 2018{\natexlab{b}}.
\newblock Robust distant supervision relation extraction via deep reinforcement
  learning.
\newblock In \emph{ACL}.

\bibitem[{Qiu et~al.(2019)Qiu, Zhang, Feng, Liao, Jiang, Lyu, Liu, and
  Zhao}]{qiu2019machine}
Delai Qiu, Yuanzhe Zhang, Xinwei Feng, Xiangwen Liao, Wenbin Jiang, Yajuan Lyu,
  Kang Liu, and Jun Zhao. 2019.
\newblock Machine reading comprehension using structural knowledge graph-aware
  network.
\newblock In \emph{EMNLP and IJCNLP}.

\bibitem[{Quirk and Poon(2017)}]{Quirk2017DistantSF}
Chris Quirk and Hoifung Poon. 2017.
\newblock Distant supervision for relation extraction beyond the sentence
  boundary.
\newblock In \emph{EACL}.

\bibitem[{Ren et~al.(2017)Ren, Wu, He, Qu, Voss, Ji, Abdelzaher, and
  Han}]{ren2017cotype}
Xiang Ren, Zeqiu Wu, Wenqi He, Meng Qu, Clare~R Voss, Heng Ji, Tarek~F
  Abdelzaher, and Jiawei Han. 2017.
\newblock {CoType}: Joint extraction of typed entities and relations with
  knowledge bases.
\newblock In \emph{WWW}.

\bibitem[{Riedel et~al.(2010)Riedel, Yao, and McCallum}]{riedel2010modeling}
Sebastian Riedel, Limin Yao, and Andrew McCallum. 2010.
\newblock Modeling relations and their mentions without labeled text.
\newblock In \emph{ECML and KDD}.

\bibitem[{Rink and Harabagiu(2010)}]{rink2010utd}
Bryan Rink and Sanda Harabagiu. 2010.
\newblock {UTD}: Classifying semantic relations by combining lexical and
  semantic resources.
\newblock In \emph{Proceedings of the 5th International Workshop on Semantic
  Evaluation}.

\bibitem[{Roth and Yih(2004)}]{roth2004linear}
Dan Roth and Wen-tau Yih. 2004.
\newblock A linear programming formulation for global inference in natural
  language tasks.
\newblock In \emph{CoNLL}.

\bibitem[{Sahu et~al.(2019)Sahu, Christopoulou, Miwa, and
  Ananiadou}]{sahu2019inter}
Sunil~Kumar Sahu, Fenia Christopoulou, Makoto Miwa, and Sophia Ananiadou. 2019.
\newblock Inter-sentence relation extraction with document-level graph
  convolutional neural network.
\newblock In \emph{ACL}.

\bibitem[{Seo et~al.(2017)Seo, Kembhavi, Farhadi, and
  Hajishirzi}]{seo2016bidirectional}
Minjoon Seo, Aniruddha Kembhavi, Ali Farhadi, and Hannaneh Hajishirzi. 2017.
\newblock Bidirectional attention flow for machine comprehension.
\newblock In \emph{ICLR}.

\bibitem[{Shang et~al.(2020)Shang, Huang, Mao, Sun, and Wei}]{Shang2020AreNS}
Yuming Shang, He-Yan Huang, Xian-Ling Mao, Xin Sun, and Wei Wei. 2020.
\newblock Are noisy sentences useless for distant supervised relation
  extraction?
\newblock In \emph{AAAI}.

\bibitem[{Shen and Huang(2016)}]{huang2016attention}
Yatian Shen and Xuanjing Huang. 2016.
\newblock Attention-based convolutional neural network for semantic relation
  extraction.
\newblock In \emph{COLING}.

\bibitem[{Shen et~al.(2019)Shen, Tan, Sordoni, and
  Courville}]{Shen2019OrderedNI}
Yikang Shen, Shawn Tan, Alessandro Sordoni, and Aaron~C. Courville. 2019.
\newblock Ordered neurons: Integrating tree structures into recurrent neural
  networks.
\newblock In \emph{ICLR}.

\bibitem[{Shi et~al.(2019)Shi, Xiao, and Niu}]{Shi2019ABS}
Y.~Shi, Y.~Xiao, and L.~Niu. 2019.
\newblock A brief survey of relation extraction based on distant supervision.
\newblock In \emph{ICCS}.

\bibitem[{Song et~al.(2018)Song, Zhang, Wang, and Gildea}]{Song2018NaryRE}
Linfeng Song, Yue Zhang, Zhiguo Wang, and Daniel Gildea. 2018.
\newblock N-ary relation extraction using graph state {LSTM}.
\newblock In \emph{EMNLP}.

\bibitem[{Sui et~al.(2021)Sui, Chen, Liu, Zhao, Zeng, and Liu}]{sui2020jointea}
Dianbo Sui, Yubo Chen, Kang Liu, Jun Zhao, Xiangrong Zeng, and Shengping Liu.
  2021.
\newblock Joint entity and relation extraction with set prediction networks.
\newblock In \emph{AAAI}.

\bibitem[{Sun et~al.(2020)Sun, Zhang, Mensah, yi~Mao, and
  Liu}]{Sun2020RecurrentIN}
Kai Sun, Richong Zhang, Samuel Mensah, Yong yi~Mao, and Xudong Liu. 2020.
\newblock Recurrent interaction network for jointly extracting entities and
  classifying relations.
\newblock In \emph{EMNLP}.

\bibitem[{Surdeanu et~al.(2012)Surdeanu, Tibshirani, Nallapati, and
  Manning}]{mimlre}
Mihai Surdeanu, Julie Tibshirani, Ramesh Nallapati, and Christopher~D. Manning.
  2012.
\newblock Multi-instance multi-label learning for relation extraction.
\newblock In \emph{EMNLP and CoNLL}.

\bibitem[{Tai et~al.(2015)Tai, Socher, and Manning}]{tai2015improved}
Kai~Sheng Tai, Richard Socher, and Christopher~D. Manning. 2015.
\newblock Improved semantic representations from tree-structured long
  short-term memory networks.
\newblock In \emph{ACL and IJCNLP}.

\bibitem[{{Takanobu} et~al.(2019){Takanobu}, {Zhang}, {Liu}, and
  {Huang}}]{takanobu2019hrlre}
Ryuichi {Takanobu}, Tianyang {Zhang}, Jiexi {Liu}, and Minlie {Huang}. 2019.
\newblock A hierarchical framework for relation extraction with reinforcement
  learning.
\newblock In \emph{AAAI}.

\bibitem[{Tang et~al.(2020)Tang, Cao, Zhang, Cao, Fang, Wang, and
  Yin}]{Tang2020HINHI}
Hengzhu Tang, Yanan Cao, Zhenyu Zhang, Jiangxia Cao, Fang Fang, Shigang Wang,
  and Pengfei Yin. 2020.
\newblock {HIN}: {H}ierarchical inference network for document-level relation
  extraction.
\newblock \emph{Advances in Knowledge Discovery and Data Mining}.

\bibitem[{Thillaisundaram and Togia(2019)}]{Thillaisundaram2019BiomedicalRE}
Ashok Thillaisundaram and Theodosia Togia. 2019.
\newblock Biomedical relation extraction with pre-trained language
  representations and minimal task-specific architecture.
\newblock \emph{ArXiv}.

\bibitem[{Trisedya et~al.(2019)Trisedya, Weikum, Qi, and
  Zhang}]{trisedya2019neural}
Bayu~Distiawan Trisedya, Gerhard Weikum, Jianzhong Qi, and Rui Zhang. 2019.
\newblock Neural relation extraction for knowledge base enrichment.
\newblock In \emph{ACL}.

\bibitem[{Turian et~al.(2010)Turian, Ratinov, and Bengio}]{turian2010word}
Joseph Turian, Lev Ratinov, and Yoshua Bengio. 2010.
\newblock Word representations: {A} simple and general method for
  semi-supervised learning.
\newblock In \emph{ACL}.

\bibitem[{Vashishth et~al.(2018)Vashishth, Joshi, Prayaga, Bhattacharyya, and
  Talukdar}]{vashishth2018reside}
Shikhar Vashishth, Rishabh Joshi, Sai~Suman Prayaga, Chiranjib Bhattacharyya,
  and Partha Talukdar. 2018.
\newblock {RESIDE}: Improving distantly-supervised neural relation extraction
  using side information.
\newblock In \emph{EMNLP}.

\bibitem[{Vaswani et~al.(2017)Vaswani, Shazeer, Parmar, Uszkoreit, Jones,
  Gomez, Kaiser, and Polosukhin}]{vaswani2017attention}
Ashish Vaswani, Noam Shazeer, Niki Parmar, Jakob Uszkoreit, Llion Jones,
  Aidan~N Gomez, {\L}ukasz Kaiser, and Illia Polosukhin. 2017.
\newblock Attention is all you need.
\newblock In \emph{NIPS}.

\bibitem[{Vela and Declerck(2009)}]{Vela2009ConceptAR}
M.~Vela and Thierry Declerck. 2009.
\newblock Concept and relation extraction in the finance domain.
\newblock In \emph{IWCS}.

\bibitem[{Veličković et~al.(2018)Veličković, Cucurull, Casanova, Romero,
  Liò, and Bengio}]{velickovic2018graph}
Petar Veličković, Guillem Cucurull, Arantxa Casanova, Adriana Romero, Pietro
  Liò, and Yoshua Bengio. 2018.
\newblock Graph attention networks.
\newblock In \emph{ICLR}.

\bibitem[{Veyseh et~al.(2020)Veyseh, Dernoncourt, Dou, and
  Nguyen}]{Veyseh2020ExploitingTS}
Amir Pouran~Ben Veyseh, Franck Dernoncourt, D.~Dou, and T.~Nguyen. 2020.
\newblock Exploiting the syntax-model consistency for neural relation
  extraction.
\newblock In \emph{ACL}.

\bibitem[{Vrandečić and Krötzsch(2014)}]{wikidata}
Denny Vrandečić and Markus Krötzsch. 2014.
\newblock Wikidata: A free collaborative knowledge base.
\newblock \emph{Communications of the ACM}.

\bibitem[{Walker et~al.(2006)Walker, Strassel, Medero, and
  Maeda}]{ace2005walker}
Christopher Walker, Stephanie Strassel, Julie Medero, and Kazuaki Maeda. 2006.
\newblock Ace 2005 multilingual training corpus.
\newblock In \emph{Linguistic Data Consortium}.

\bibitem[{Wang et~al.(2020{\natexlab{a}})Wang, Hu, Cao, and
  Sun}]{Wang2020GlobaltoLocalNN}
D.~Wang, Wei Hu, E.~Cao, and Weijian Sun. 2020{\natexlab{a}}.
\newblock Global-to-local neural networks for document-level relation
  extraction.
\newblock In \emph{EMNLP}.

\bibitem[{Wang et~al.(2019)Wang, Focke, Sylvester, Mishra, and
  Wang}]{Wang2019FinetuneBF}
Hong Wang, Christfried Focke, Rob Sylvester, Nilesh Mishra, and William W.~J.
  Wang. 2019.
\newblock Fine-tune {BERT} for {DocRED} with two-step process.
\newblock \emph{ArXiv}.

\bibitem[{Wang and Lu(2020)}]{Wang2020TwoAB}
Jue Wang and Wei Lu. 2020.
\newblock Two are better than one: Joint entity and relation extraction with
  table-sequence encoders.
\newblock In \emph{EMNLP}.

\bibitem[{Wang et~al.(2016)Wang, Cao, de~Melo, and Liu}]{wang2016relation}
Linlin Wang, Zhu Cao, Gerard de~Melo, and Zhiyuan Liu. 2016.
\newblock Relation classification via multi-level attention {CNN}s.
\newblock In \emph{ACL}.

\bibitem[{Wang et~al.(2020{\natexlab{b}})Wang, Yu, Zhang, Liu, Zhu, and
  Sun}]{wang2020tplinker}
Yucheng Wang, Bowen Yu, Y.~Zhang, Tingwen Liu, Hongsong Zhu, and L.~Sun.
  2020{\natexlab{b}}.
\newblock Tplinker: Single-stage joint extraction of entities and relations
  through token pair linking.
\newblock In \emph{COLING}.

\bibitem[{Wei et~al.(2020)Wei, Su, Wang, Tian, and Chang}]{Wei2020ANC}
Zhepei Wei, Jianlin Su, Yue Wang, Yuan Tian, and Yi~Chang. 2020.
\newblock A novel cascade binary tagging framework for relational triple
  extraction.
\newblock In \emph{ACL}.

\bibitem[{Welbl et~al.(2018)Welbl, Stenetorp, and
  Riedel}]{welbl2018constructing}
Johannes Welbl, Pontus Stenetorp, and Sebastian Riedel. 2018.
\newblock Constructing datasets for multi-hop reading comprehension across
  documents.
\newblock In \emph{TACL}.

\bibitem[{Wu et~al.(2019)Wu, Fan, and Zhang}]{wu2018improving}
Shanchan Wu, Kai Fan, and Qiong Zhang. 2019.
\newblock Improving distantly supervised relation extraction with neural noise
  converter and conditional optimal selector.
\newblock In \emph{AAAI}.

\bibitem[{Wu et~al.(2017)Wu, Bamman, and Russell}]{wu2017adversarial}
Yi~Wu, David Bamman, and Stuart Russell. 2017.
\newblock Adversarial training for relation extraction.
\newblock In \emph{EMNLP}.

\bibitem[{Xu et~al.(2015)Xu, Mou, Li, Chen, Peng, and Jin}]{xu2015classifying}
Yuning Xu, Lili Mou, Ge~Li, Yunchuan Chen, Hao Peng, and Zhi Jin. 2015.
\newblock Classifying relations via long short term memory networks along
  shortest dependency paths.
\newblock In \emph{EMNLP}.

\bibitem[{Yaghoobzadeh et~al.(2017)Yaghoobzadeh, Adel, and
  Sch{\"u}tze}]{yaghoobzadeh2017noise}
Yadollah Yaghoobzadeh, Heike Adel, and Hinrich Sch{\"u}tze. 2017.
\newblock Noise mitigation for neural entity typing and relation extraction.
\newblock In \emph{EACL}.

\bibitem[{Yao et~al.(2019)Yao, Ye, Li, Han, Lin, Liu, Liu, Huang, Zhou, and
  Sun}]{yao2019DocRED}
Yuan Yao, Deming Ye, Peng Li, Xu~Han, Yankai Lin, Zhenghao Liu, Zhiyuan Liu,
  Lixin Huang, Jie Zhou, and Maosong Sun. 2019.
\newblock {DocRED}: A large-scale document-level relation extraction dataset.
\newblock In \emph{ACL}.

\bibitem[{Yates et~al.(2007)Yates, Cafarella, Banko, Etzioni, Broadhead, and
  Soderland}]{yates2007textrunner}
Alexander Yates, Michael Cafarella, Michele Banko, Oren Etzioni, Matthew
  Broadhead, and Stephen Soderland. 2007.
\newblock {TEXTRUNNER}: {O}pen information extraction on the web.
\newblock In \emph{NAACL-HLT}.

\bibitem[{Ye et~al.(2021)Ye, Zhang, Deng, Chen, Tan, Huang, and
  Chen}]{ye2020contrastive}
Hongbin Ye, Ningyu Zhang, Shumin Deng, M.~Chen, Chuanqi Tan, Fei Huang, and
  Huajun Chen. 2021.
\newblock Contrastive triple extraction with generative transformer.
\newblock \emph{AAAI}.

\bibitem[{Ye and Ling(2019)}]{ye2019distant}
Zhi-Xiu Ye and Zhen-Hua Ling. 2019.
\newblock Distant supervision relation extraction with intra-bag and inter-bag
  attentions.
\newblock In \emph{NAACL-HLT}.

\bibitem[{Yuan et~al.(2019)Yuan, Liu, Tang, Zhang, Zhuang, Pu, Wu, and
  Ren}]{Yuan2019CrossrelationCA}
Y.~Yuan, Liyuan Liu, Siliang Tang, Zhongfei Zhang, Y.~Zhuang, S.~Pu, Fei Wu,
  and Xiang Ren. 2019.
\newblock Cross-relation cross-bag attention for distantly-supervised relation
  extraction.
\newblock In \emph{AAAI}.

\bibitem[{Yuan et~al.(2020)Yuan, Zhou, Pan, Zhu, Song, and Guo}]{Yuan2020ARA}
Yue Yuan, Xiaofei Zhou, Shirui Pan, Qiannan Zhu, Zeliang Song, and Li~Guo.
  2020.
\newblock A relation-specific attention network for joint entity and relation
  extraction.
\newblock In \emph{IJCAI}.

\bibitem[{Zeng et~al.(2015)Zeng, Liu, Chen, and Zhao}]{zeng2015distant}
Daojian Zeng, Kang Liu, Yubo Chen, and Jun Zhao. 2015.
\newblock Distant supervision for relation extraction via piecewise
  convolutional neural networks.
\newblock In \emph{EMNLP}.

\bibitem[{Zeng et~al.(2014)Zeng, Liu, Lai, Zhou, and Zhao}]{zeng2014relation}
Daojian Zeng, Kang Liu, Siwei Lai, Guangyou Zhou, and Jun Zhao. 2014.
\newblock Relation classification via convolutional deep neural network.
\newblock In \emph{COLING}.

\bibitem[{Zeng et~al.(2020{\natexlab{a}})Zeng, Zhang, and
  Liu}]{Zeng2020CopyMTLCM}
Daojian Zeng, Haoran Zhang, and Qianying Liu. 2020{\natexlab{a}}.
\newblock {CopyMTL}: {C}opy mechanism for joint extraction of entities and
  relations with multi-task learning.
\newblock In \emph{AAAI}.

\bibitem[{Zeng et~al.(2020{\natexlab{b}})Zeng, Xu, Chang, and
  Li}]{Zeng2020DoubleGB}
Shuang Zeng, Runxin Xu, Baobao Chang, and Lei Li. 2020{\natexlab{b}}.
\newblock Double graph based reasoning for document-level relation extraction.
\newblock In \emph{ACL}.

\bibitem[{Zeng et~al.(2018)Zeng, Zeng, He, Liu, and Zhao}]{zeng2018copyre}
Xiangrong Zeng, Daojian Zeng, Shizhu He, Kang Liu, and Jun Zhao. 2018.
\newblock Extracting relational facts by an end-to-end neural model with copy
  mechanism.
\newblock In \emph{ACL}.

\bibitem[{Zhang et~al.(2019)Zhang, Li, Jia, and Hai}]{Zhang2019MultilabeledRE}
Xinsong Zhang, P.~Li, W.~Jia, and Zhao Hai. 2019.
\newblock Multi-labeled relation extraction with attentive capsule network.
\newblock In \emph{AAAI}.

\bibitem[{Zhang et~al.(2018)Zhang, Qi, and Manning}]{zhang2018graph}
Yuhao Zhang, Peng Qi, and Christopher~D. Manning. 2018.
\newblock Graph convolution over pruned dependency trees improves relation
  extraction.
\newblock In \emph{EMNLP}.

\bibitem[{Zhang et~al.(2017)Zhang, Zhong, Chen, Angeli, and
  Manning}]{zhang2017position}
Yuhao Zhang, Victor Zhong, Danqi Chen, Gabor Angeli, and Christopher~D.
  Manning. 2017.
\newblock Position-aware attention and supervised data improve slot filling.
\newblock In \emph{EMNLP}.

\bibitem[{Zhao et~al.(2020)Zhao, Zhang, qing Zhou, and
  Zong}]{Zhao2020KnowledgeGE}
Yang Zhao, Jiajun Zhang, Yin qing Zhou, and Chengqing Zong. 2020.
\newblock Knowledge graphs enhanced neural machine translation.
\newblock In \emph{IJCAI}.

\bibitem[{Zheng et~al.(2017)Zheng, Wang, Bao, Hao, Zhou, and
  Xu}]{zheng2017joint}
Suncong Zheng, Feng Wang, Hongyun Bao, Yuexing Hao, Peng Zhou, and Bo~Xu. 2017.
\newblock Joint extraction of entities and relations based on a novel tagging
  scheme.
\newblock In \emph{ACL}.

\bibitem[{Zhou et~al.(2020)Zhou, Xu, Yao, Liu, Lang, and
  Jiang}]{Zhou2020GlobalCG}
Huiwei Zhou, Yibin Xu, W.~Yao, Zhe Liu, Chengkun Lang, and H.~Jiang. 2020.
\newblock Global context-enhanced graph convolutional networks for
  document-level relation extraction.
\newblock In \emph{COLING}.

\bibitem[{Zhu et~al.(2020)Zhu, Wang, Yu, Zhou, Chen, Zhang, and
  Zhang}]{zhu2020nyth}
Tong Zhu, Haitao Wang, Junjie Yu, Xiabing Zhou, Wenliang Chen, Wei Zhang, and
  Min Zhang. 2020.
\newblock Towards accurate and consistent evaluation: A dataset for
  distantly-supervised relation extraction.
\newblock In \emph{COLING}.

\end{thebibliography}

\end{document}